\documentclass[preprint,11pt,number]{elsarticle}

\usepackage{graphicx}
\usepackage{amsmath}
\usepackage{amssymb}
\usepackage{bm}
\usepackage{makecell}
\usepackage{multirow}
\usepackage[table]{xcolor}
\usepackage{float}
\usepackage[section]{placeins}
\usepackage{paralist}
\usepackage{microtype}
\usepackage{algorithm}
\usepackage{algpseudocode}
\usepackage{url}
\usepackage{orcidlink}
\usepackage[labelfont=bf]{caption}
\usepackage{booktabs}
\usepackage{subfigure}
\usepackage{subcaption}
\usepackage{siunitx}

\journal{preprint}

\begin{document}

\begin{frontmatter}

\title{Track--Detection Link Prediction for Learned Association in Multi-Object Tracking}

\author[ub]{Momir Adžemović\corref{cor1}\fnref{orcid1}}
\ead{pd222011@alas.matf.bg.ac.rs}

\cortext[cor1]{Corresponding author.}
\fntext[orcid1]{ORCID: 0009-0001-1385-4094}

\affiliation[ub]{
    organization={Department of Computer Science, Faculty of Mathematics, University of Belgrade},
    addressline={Studentski trg 16},
    city={Belgrade},
    postcode={11000},
    country={Serbia}
}

\begin{abstract}

Multi-object tracking aims to maintain object identities over time by associating detections across video frames. Two dominant paradigms exist in the literature: tracking-by-detection methods, which are computationally efficient but rely on handcrafted association heuristics, and end-to-end approaches, which learn associations from data at the cost of higher computational complexity. We propose Track-Detection Link Prediction (TDLP), a tracking-by-detection method that performs per-frame association via link prediction between tracks and detections, i.e., by predicting the correct continuation of each track at every frame. TDLP is architecturally designed primarily for geometric features such as bounding boxes, while optionally incorporating additional cues, including pose and appearance. Unlike heuristic-based methods, TDLP learns associations directly from data without using any domain-specific heuristics, while remaining modular and computationally efficient compared to end-to-end trackers. Extensive experiments on multiple benchmarks demonstrate that TDLP consistently surpasses state-of-the-art performance against both tracking-by-detection and end-to-end methods. Finally, we provide a detailed analysis comparing link prediction with metric learning-based association and show that link prediction is more effective, particularly when handling heterogeneous features such as detection bounding boxes. Our code is available at \href{https://github.com/Robotmurlock/TDLP}{https://github.com/Robotmurlock/TDLP}.

\end{abstract}


\end{frontmatter}

\section{Introduction}
\label{sec:introduction}

Multi-object tracking (MOT) aims to localize and maintain the identities of multiple objects over time and is a core component of many video-based perception systems. It is important in a wide range of applications, including autonomous driving~\cite{nuScenes,simtrack}, sports analytics~\cite{sportsmot,soccernet}, retail analytics~\cite{retail_analytics}, robotics~\cite{object_tracking_in_robotics}, and surveillance~\cite{object_tracking_in_surveillence}. Despite significant advances in object detection, MOT remains challenging: occlusions, crowded scenes, camera motion, and highly dynamic, non-linear trajectories place strong demands on both accuracy and efficiency.

Many modern MOT approaches adopt the tracking-by-detection paradigm, in which object detections are obtained independently in each frame and subsequently associated across time~\cite{bytetrack,botsort,ocsort,hybridsort,mot_survey2025}. Early methods such as SORT~\cite{sort} and DeepSORT~\cite{deepsort} formalized this as a per-frame bipartite matching problem between existing tracks and detections in the current frame . These approaches usually rely on Kalman filtering for motion prediction and optionally use appearance features~\cite{bytetrack,botsort,hybridsort,ocsort,deepocsort,boostrack,sparsetrack,deep_eiou}. Their performance also depends on carefully designed heuristics. While such trackers are computationally efficient, their reliance on domain-specific assumptions limits their ability to generalize across datasets. At the other end of the spectrum, fully end-to-end MOT methods learn detection and association jointly from data~\cite{motr,memotr,motip,mot_survey2025} and therefore require no handcrafted association rules. However, they usually require large training datasets and incur high computational costs. As a result, they are often too slow for real-time use~\cite{motr,memotr,motip}. This creates a gap between efficient heuristic-based trackers and expressive but computationally expensive end-to-end methods.

\begin{figure}
  \centering
  \includegraphics[width=1.0\linewidth]{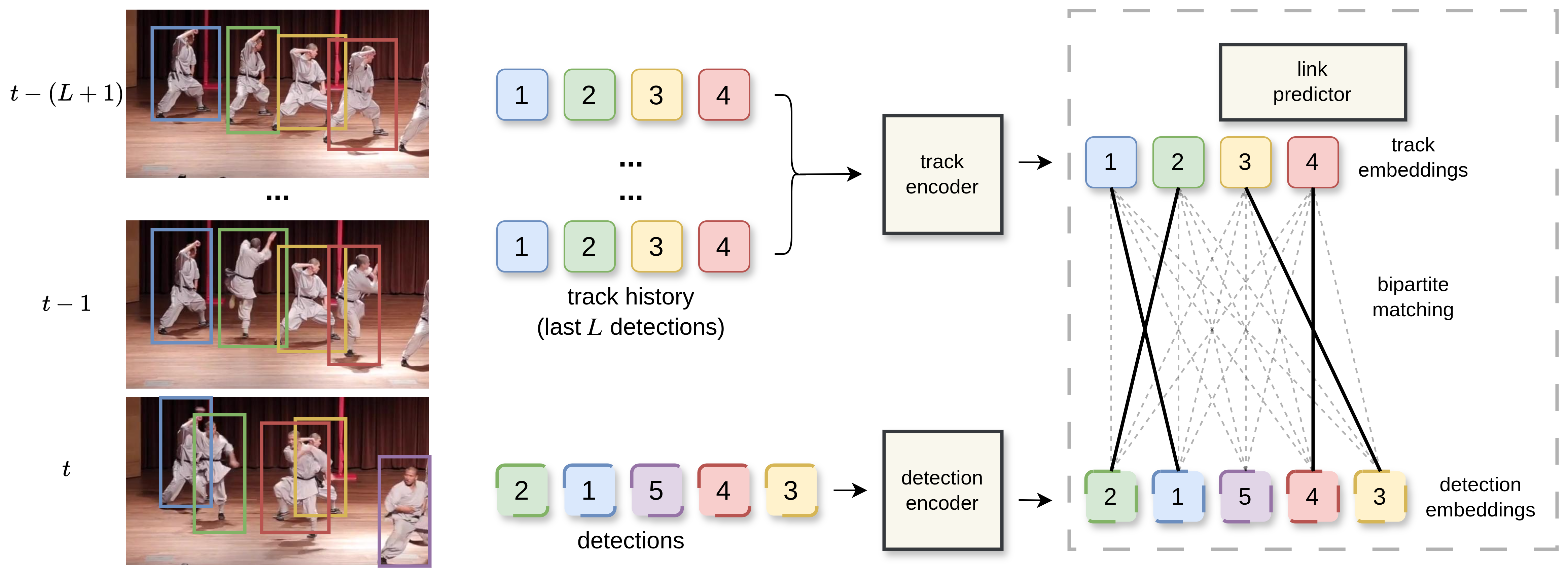}
  \caption{High-level overview of the proposed association pipeline. Encoded track histories and current detections are linked via link prediction and associated using bipartite matching.}
  \label{fig:high_level}
\end{figure}

In this work, we propose Track–Detection Link Prediction (TDLP), a tracking-by-detection method that lies between heuristic-based association and fully end-to-end learning. TDLP formulates data association as a per-frame link prediction problem between existing tracks and current frame detections. Track and detection features are first embedded into compact vector representations, and link probabilities are then predicted between these vectors. A high-level overview is shown in Figure~\ref{fig:high_level}. TDLP learns associations directly from data without using any domain-specific heuristics and maintains a lower computational cost than end-to-end methods during both training and inference. To improve tracking accuracy, the method employs an architecture designed to model non-linear patterns in real-world data. It further supports feature aggregation across multiple modalities, including bounding boxes, appearance features, and human pose keypoints. Even without appearance features, TDLP outperforms heuristic-based methods on datasets characterized by non-linear motion. When additional cues are incorporated, TDLP achieves state-of-the-art performance on DanceTrack~\cite{dancetrack}, SportsMOT~\cite{sportsmot}, SoccerNet~\cite{soccernet}, and BEE24~\cite{bee24}. The contributions of this paper are as follows:

\begin{itemize}
\item We propose TDLP, a learning-based association method that performs per-frame bipartite matching via link prediction, leveraging advanced motion modeling and multi-modal feature aggregation.
\item We design a SORT-like tracking framework based on TDLP that surpasses the state-of-the-art. Notably, when using only bounding boxes provided by the detector, our method outperforms all heuristic-based trackers, including those that use object appearance cues.
\item We analyze the proposed link prediction formulation and compare it with alternative association formulations such as metric learning, identifying the key factors that cause metric learning to underperform relative to our approach.
\end{itemize}

\section{Related Work}
\label{sec:related_work}

\textbf{Heuristic-based tracking-by-detection.}
Modern tracking-by-detection methods largely build on the two-stage association strategy introduced by ByteTrack~\cite{bytetrack}, which first associates high-confidence detections before considering low-confidence ones.
Subsequent works extend this pipeline with engineered heuristics, such as exploiting perspective effects via object height or image-plane position~\cite{movesort,hybridsort,sparsetrack,boostrack}, incorporating detection confidence~\cite{boostrack,hybridsort,deepmovesort}, compensating for camera motion~\cite{botsort,deepocsort,hybridsort,ucmctrack}, or using improved motion models~\cite{ettrack,motiontrack,movesort,deepmovesort,ocsort}.
While effective, such heuristics are often dataset- and scenario-dependent and require careful hyper-parameter tuning.
In the literature, they tend to perform better on crowded datasets with relatively linear motion, such as MOTChallenge~\cite{mot_challenge}, than on datasets with non-linear motion and visually similar objects, such as DanceTrack~\cite{dancetrack} and SportsMOT~\cite{sportsmot,mot_survey2025}.
In contrast, our approach removes handcrafted association rules and learns the association function directly from data.
This enables adaptive weighting of cues based on their reliability and improves robustness on datasets with non-linear motion.

\textbf{End-to-end tracking.} End-to-end trackers jointly perform detection and association within a single model, avoiding explicit heuristics. Since MOTR~\cite{motr}, this line of work has steadily advanced and achieves strong performance on challenging benchmarks such as DanceTrack~\cite{memotr,motip,dancetrack}. MOTIP~\cite{motip} takes a different direction by formulating online MOT as in-context ID prediction, where historical trajectories provide identity prompts and current detections are classified into the corresponding prompted IDs.
However, MOTIP is still built around a DETR-style detector and relies on object-level detector embeddings for association, making its computational profile closer to that of end-to-end transformer trackers. End-to-end methods are computationally expensive to train and deploy~\cite{motr,motip,memot}, and their tightly coupled design reduces modularity and complicates component-level modifications. Our method offers a simpler alternative that retains the benefits of learned association while preserving the efficiency and modularity of tracking-by-detection pipelines. By optionally incorporating appearance features, it enables a direct trade-off between accuracy and inference speed.

\textbf{Offline graph-based methods.}
Our work is related to offline graph-based trackers such as MPNTrack~\cite{mpntrack} and SUSHI~\cite{sushi}, which formulate data association as a link prediction problem over spatio-temporal graphs.
These methods perform global optimization over video clips and exploit long-range temporal context.
However, they operate offline and require access to future frames, limiting their applicability in real-time scenarios.
In contrast, our architectural design performs link prediction online by modeling associations between tracks and detections as track continuations.

\textbf{Tracking-by-detection with learned association.}
Concurrent works propose using metric learning to learn associations between tracks and detections, where embeddings are matched using distance metrics such as cosine or normalized Euclidean distance~\cite{cameltrack,twix}.
However, metric learning enforces a global embedding structure, which can be restrictive when handling heterogeneous features, particularly low-dimensional geometric cues such as bounding boxes.
In such settings, embeddings may remain close unless differences accumulate across multiple feature dimensions, leading to incorrect associations even when a mismatch in a single cue should be sufficient.
In contrast, the link prediction model outputs similarity at the feature level and can emphasize the most informative differences between objects, making it particularly suitable for geometric association.
We analyze this distinction in Section~\ref{sec:ablation_studies} and show that link prediction consistently outperforms contrastive learning when using bounding box features.

PuTR~\cite{putr} is also related to this line of work, as it separates detection and association and uses a pure Transformer to model a temporally ordered object sequence.
However, PuTR relies on appearance features extracted from detected objects and performs sequence-level association, which introduces additional computational cost.
In contrast, TDLP performs per-frame track-detection link prediction and can operate using only detection bounding boxes.
Despite using only geometric information, TDLP achieves stronger performance than PuTR in our experiments while running faster.
Moreover, TDLP is feature-agnostic: the same link prediction pipeline can incorporate additional cues, such as appearance or pose features, when they are available.

\section{Methodology}
\label{sec:methodology}

We formulate per-frame association as a bipartite link prediction problem between tracks and detections, define the corresponding optimization objective, and present the TDLP architecture and tracking pipeline used for frame-wise association.

\subsection{Track--detection bipartite link prediction}
\label{sec:bipartite_link_prediction}

We consider an online MOT setting where, at each frame $t$, we maintain a set of active tracks 
$\mathcal{T}_t = \{T_1, \dots, T_{N_t}\}$ and a set of detections 
$\mathcal{D}_t = \{D_1, \dots, D_{M_t}\}$ produced by an object detector. Each track $T_i$ is represented by a short temporal window of length $L$ containing its most recent observations. Our objective is to estimate the association matrix 
$\mathbf{Y}_t \in \{0,1\}^{N_t \times M_t}$, where $Y_{ij} = 1$ if detection $D_j$ is the correct continuation of track $T_i$ at frame $t$, and $Y_{ij} = 0$ otherwise. In other words, we aim to approximate the adjacency matrix of the bipartite graph formed between tracks and detections.

Our proposed link prediction pipeline proceeds as follows. Let $f_{t}$ and $f_{d}$ denote the track and detection encoder functions, respectively. These functions transform the raw input features into a unified embedding space. For each track $T_i$ and detection $D_j$, the model computes
\begin{equation}
\label{eq:track_det_encodings}
\bm{e}^{\mathrm{trk}}_{i} = f_{t}(T_i), \qquad 
\bm{e}^{\mathrm{det}}_{j} = f_{d}(D_j).
\end{equation}
Here, $\bm{e}^{\mathrm{trk}}_{i}$ and $\bm{e}^{\mathrm{det}}_{j}$ are the encoded representations produced by the architecture described in Section~\ref{sec:architecture}. The link probability for each pair $(i,j)$ is then predicted as
\begin{equation}
\label{eq:similarity_score}
S_{ij} = \phi\!\left(\bm{e}^{\mathrm{trk}}_{i},\, \bm{e}^{\mathrm{det}}_{j}\right),
\end{equation}
where $\phi$ denotes the track--detection link prediction function. The resulting score $S_{ij} \in (0,1)$ is interpreted as the probability that track $T_i$ and detection $D_j$ correspond to the same trajectory and should therefore be linked.

In order to learn $f_{t}$, $f_{d}$, and $\phi$, we sample video clips from the dataset and optimize the weighted binary cross-entropy (BCE) loss over all positive and negative track--detection pairs:
\begin{equation}
    \mathcal{L}_{\text{BCE}} 
    = - \sum_{i=1}^{N_t} \sum_{j=1}^{M_t}
    \Big( 
        w^{+} Y_{ij} \log S_{ij} 
        + (1 - Y_{ij}) \log (1 - S_{ij}) 
    \Big),
    \label{eq:tdlp_bce_loss}
\end{equation}
where $Y_{ij}$ is derived from ground-truth identities. The parameter $w^{+}>1$ controls the positive-class weighting and compensates for the strong imbalance between positive and negative samples, i.e., the fact that most track--detection pairs correspond to negative links. Only the final clip frame contributes to the supervision.

At inference time, the scores $S_{ij}$ define the edge weights in a bipartite matching between tracks and detections, and a simple linear assignment (e.g. the Hungarian algorithm)~\cite{linear_assignment} is applied to obtain the predicted associations.

\subsection{Architecture}
\label{sec:architecture}

\begin{figure}
  \centering
  \includegraphics[width=1.0\linewidth]{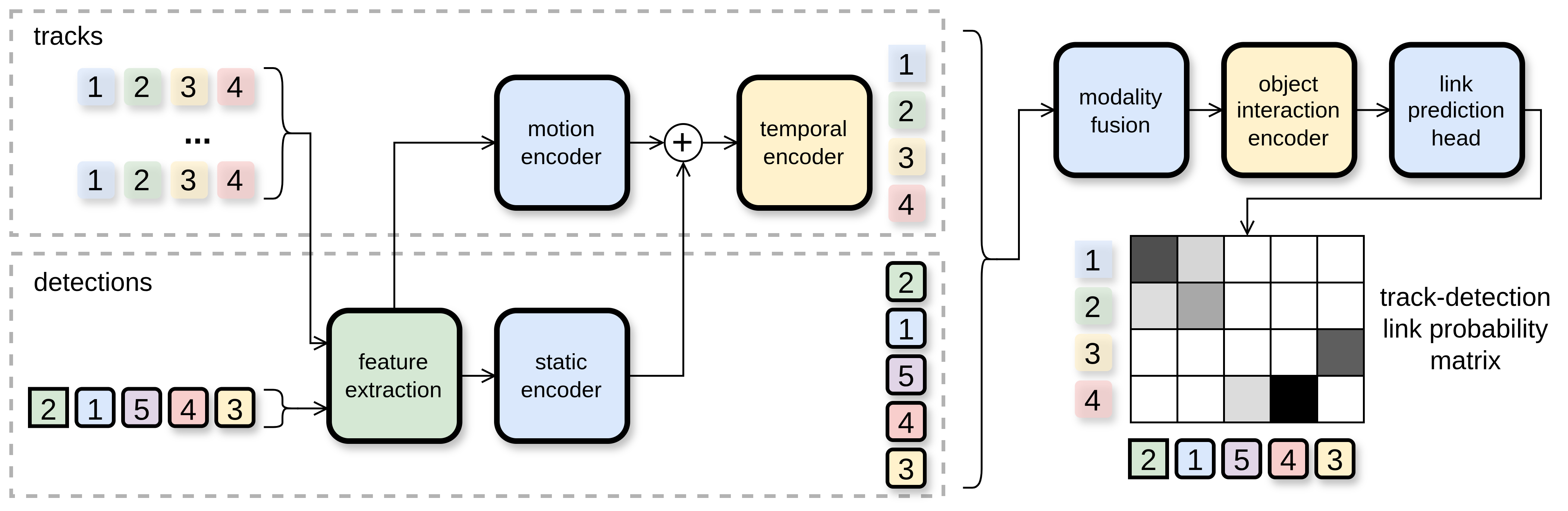}
  \caption{Overview of the proposed TDLP architecture.
Tracks and detections first undergo a geometric feature transformation (GFT), while optional additional cues (e.g., pose keypoints and appearance) are extracted using dedicated models. The static encoder processes per-frame detections from both current detections and track histories, while a motion encoder extracts track-specific temporal features (e.g., velocity). Track features are aggregated over time using a transformer-based temporal encoder. The resulting track and detection embeddings are fused across modalities and refined by an transformer-based object-interaction encoder. Finally, a link prediction head outputs association probabilities for each track–detection pair.}
  \label{fig:architecture}
\end{figure}

The TDLP architecture (Figure~\ref{fig:architecture}) follows a transformer-encoder design that maps tracks and detections into representations suitable for bipartite link prediction. The baseline variant, \textit{TDLP-bbox}, uses only bounding box geometry and confidence scores, while additional modalities (e.g., appearance) can be incorporated to improve accuracy. The pipeline consists of three stages: (i) feature extraction, (ii) feature encoding, and (iii) link prediction.

\textbf{Feature extraction}. Prior work shows that simple geometric transformations can improve tracking accuracy~\cite{movesort,deepmovesort,motiontrack,ettrack}. Following this observation, we adopt a feature extraction pipeline tailored to link prediction. At time $t$, the model processes up to $N_t \cdot L + M_t$ detections, where $N_t$ is the number of active tracks, $L$ is the history length, and $M_t$ is the number of current detections. Let $x_{i,t}$ denote the detection features of object $i$ at time $t$.

To reduce sensitivity to absolute image coordinates and improve generalization across clips, we apply \emph{frame-wise min–max normalization} to all geometric features, mapping each coordinate to $[0,1]$:
\begin{equation}
\tilde{x} = \frac{x - x_{\min}}{x_{\max} - x_{\min}}.
\end{equation}
To capture rough motion patterns, we compute \emph{first-order differences} for each track:
\begin{equation}
\Delta x_{i,t} = \frac{x_{i,t} - x_{i,\tau}}{t - \tau},
\end{equation}
where $\tau$ denotes the time index of the most recent observation of object~$i$. This operation is applied only to tracks, as detections lack a temporal dimension and therefore do not support temporal differences. Finally, all feature dimensions are \emph{standardized} to zero mean and unit variance. These transformations are general and domain-agnostic, and naturally extend to additional geometric modalities such as human pose keypoints (hereafter referred to as \emph{keypoints}).

Track and detection inputs cannot be processed by a shared encoder because their feature sets may differ (e.g., velocity estimates are defined only for tracks). Accordingly, we use a static encoder $f_{\text{stat}}$ for all point observations and a motion encoder $f_{\text{mot}}$ for track-specific features. The resulting input embeddings are
\begin{equation}
    \bm{h}_{i,t-k} = f_{\text{stat}}(x_{i,t-k}) + f_{\text{mot}}(\Delta x_{i,t-k}),
    \qquad
    \bm{h}_{j,t} = f_{\text{stat}}(x_{j,t}),
    \label{eq:tdlp_feature_encoding}
\end{equation}
where $k \in \{1,2,\dots,L\}$.

Appearance features extracted by re-identification (ReID) models are commonly used to improve association performance~\cite{botsort,deep_eiou,deepocsort,deepmovesort,hybridsort}. Thus, our architecture supports the inclusion of additional modalities that can be derived from detections (e.g., appearance, keypoints, or other object attributes). In the remainder of the text, we use $x^{(m)}_{i,t}$ to denote a feature vector of modality $m$ for the $i$-th object at time $t$.

\textbf{Temporal transformer encoder}. For temporal track modeling, we adopt the motion encoder from TransFilter~\cite{deepmovesort}, a transformer architecture that is robust to trajectory truncation via reversed positional encoding (RPE). Since our objective is to obtain a track embedding for the next frame rather than predict a bounding box, we remove the prediction head and use the transformer output directly:
\begin{equation}
    \bm{z}^{trk}_{i} = f_{\text{temporal}}\!\left(\bm{h}_{i,t-L:t-1}\right),
    \label{eq:tdlp_temporal_encoder}
\end{equation}
where $f_{\text{temporal}}$ denotes the temporal transformer encoder and $\bm{h}_{i,t-L:t-1}$ is the sequence of input embeddings for track $i$. For detections, $\bm{z}^{det}_{j} = \bm{h}_{j,t}$.

\textbf{Multi-modal fusion.} After track temporal encoding, each object yields modality-specific tokens ${\mathbf{z}^{(m)}}_{m\in\mathcal{M}}$. Inspired by CAMELTrack~\cite{cameltrack}, we map each modality into a shared embedding space via a learned linear projection $\mathbf{W}^{(m)}$, and the final multi-modal token is obtained by summing the projected embeddings:

\begin{equation}
\mathbf{u}
= \sum_{m \in \mathcal{M}} \mathbf{W}^{(m)} \mathbf{z}^{(m)}.
\label{eq:tdlp_mm_fusion}
\end{equation}

\textbf{Object interaction encoder}. To model interactions between objects within the same frame, we apply transformer-style encoders to both tracks and detections. A single-modality interaction encoder is first applied independently to each modality (omitted in Figure~\ref{fig:architecture} to avoid clutter). After multi-modal fusion, a joint interaction encoder processes all tokens:
\begin{equation}
    \big\{ \bar{\bm{z}}^{\mathrm{trk}}_i \big\}_i,\;
    \big\{ \bar{\bm{z}}^{\mathrm{det}}_j \big\}_j
    = f_{\mathrm{inter}}\!\left(
        \{\bm{u}^{\mathrm{trk}}_i\}_i \cup
        \{\bm{u}^{\mathrm{det}}_j\}_j
    \right),
\end{equation}
where $f_{\mathrm{inter}}$ is a transformer encoder applied over the combined set of multi-modal track and detection tokens. This stage allows tracks and detections to attend to each other and refine their representations before link prediction.

\textbf{Bipartite link prediction head}. Given refined track and detection embeddings
$\bar{\bm{z}}^{\text{trk}}_i$ and $\bar{\bm{z}}^{\text{det}}_j$, TDLP performs link prediction on the track--detection bipartite graph. For each pair $(i,j)$, we construct a pairwise representation
\begin{equation}
    \bm{v}_{ij} =
    \big[ \bar{\bm{z}}^{\text{trk}}_i,\;
           \bar{\bm{z}}^{\text{det}}_j,\;
           \big| \bar{\bm{z}}^{\text{trk}}_i - \bar{\bm{z}}^{\text{det}}_j \big| \big],
    \label{eq:tdlp_pairwise_representation}
\end{equation}
where $[\cdot]$ denotes the concatenation operation. A shallow MLP $\phi$ outputs the link probability $S_{ij}$ used in Eq.~\eqref{eq:tdlp_bce_loss}. At inference, the resulting scores define the cost matrix for the linear assignment task.

\subsection{The tracker}
\label{sec:tdlp_inference}

During tracking inference, TDLP follows the same high-level workflow as SORT~\cite{sort} (or DeepSORT~\cite{deepsort} when appearance features are used), but without any handcrafted association heuristics. For each frame, detections are first obtained and filtered using the detection threshold $\theta_{\mathrm{det}}$ to remove low-confidence observations. The remaining detections are optionally enriched with additional features, and TDLP computes probabilistic link scores for every track--detection pair. During association, candidate links are considered only if their predicted probability exceeds the gating threshold $\theta_{\mathrm{link}}$, after which the final one-to-one matching is computed using the Hungarian algorithm~\cite{linear_assignment}.

After association, the track life cycle is updated as follows. An unmatched detection provides evidence for initializing a new track, which becomes confirmed only after accumulating at least $T_{\mathrm{init}}$ successful associations and meeting the detection-confidence requirement $\theta_{\mathrm{new}}$. Unmatched tracks are retained for up to $T_{\mathrm{lost}}$ frames and removed if no compatible detection appears within this window.

\section{Experimental Analysis}
\label{sec:experiments}

We evaluate on five MOT datasets: DanceTrack~\cite{dancetrack}, SportsMOT~\cite{sportsmot}, SoccerNet~\cite{soccernet}, MOT17~\cite{mot_challenge}, and BEE24~\cite{bee24}. We report results for two variants: \textit{TDLP}, which combines bounding boxes, human pose keypoints, and appearance features, and the lightweight \textit{TDLP-bbox}, which uses only bounding box features. We primarily report HOTA, along with AssA and IDF1 to emphasize association performance~\cite{hota}. All implementation details are provided in Appendix~\ref{appendix:implementation}.

\subsection{Benchmarks}
\label{sec:benchmarks}

\textbf{DanceTrack.} DanceTrack\footnote{DanceTrack GitHub page: \url{https://github.com/DanceTrack/DanceTrack}.}
 is a multi-object tracking dataset of dance videos. It is particularly challenging due to visually similar performers, frequent occlusions, and fast, non-linear motion. The dataset spans diverse styles, including classical, street, pop, large-group, and sports performances, making it a demanding benchmark~\cite{dancetrack}. In the literature, the leaderboard is dominated by end-to-end trackers such as MOTIP~\cite{motip} and MeMOTR~\cite{memotr}, which typically outperform tracking-by-detection methods due to the dataset’s strong emphasis on association difficulty.

As shown in Table~\ref{tab:results_test_dancetrack}, TDLP outperforms both end-to-end and tracking-by-detection methods, improving over the previous state-of-the-art method (CAMELTrack) by $0.8\%$ HOTA and $0.9\%$ IDF1. Interestingly, the lightweight \textit{TDLP-bbox} variant surpasses all heuristic-based trackers despite relying only on bounding box features, without using appearance cues.

\begin{table*}
\scriptsize
\centering
\begin{tabular}{lccccc}
Method & HOTA & DetA & AssA & MOTA & IDF1 \\
\hline
\textit{e2e} & \multicolumn{5}{c}{} \\
MOTR~\cite{motr} & 54.2 & 73.5 & 40.2 & 79.7 & 51.5 \\
MOTIP~\cite{motip} & 67.5 & 71.6 & 57.6 & 90.3 & 72.2 \\
MeMOTR~\cite{memotr} & 68.5 & 71.6 & 58.4 & 89.9 & 71.2 \\
\hline
\textit{tbd (bbox features)} & \multicolumn{5}{c}{} \\
ByteTrack~\cite{bytetrack} & 47.3 & 71.6 & 31.4 & 89.5 & 52.5 \\
OC\_SORT~\cite{ocsort} & 55.1 & 80.4 & 38.0 & 89.4 & 54.9 \\
MoveSORT~\cite{movesort} & 56.1 & 81.6 & 38.7 & 91.8 & 56.0 \\
\textbf{TDLP-bbox (ours)} & 67.8 & \textit{82.2} & 56.1 & \textit{91.9} & 72.7 \\
\hline
\textit{tbd (extra features)} & \multicolumn{5}{c}{} \\
Deep OC\_SORT~\cite{deepocsort} & 61.3 & 81.6 & 45.8 & \textit{91.8} & 61.5 \\
DeepMoveSORT~\cite{deepmovesort} & 63.0 & 82.0 & 48.6 & \textbf{92.6} & 65.0 \\
Hybrid-SORT~\cite{hybridsort} & 65.7 & - & - & 91.8 & 67.4 \\
CAMELTrack~\cite{cameltrack} & \textit{69.3} & 81.8 & \textit{58.9} & 91.4 & \textit{74.9} \\
\textbf{TDLP (ours)} & \textbf{70.1} & \textbf{82.6} & \textbf{59.6} & 91.8 & \textbf{75.8} \\
\end{tabular}
\caption{Evaluation results on the DanceTrack test set. \textit{e2e} denotes end-to-end methods and \textit{tbd} denotes tracking-by-detection. All \textit{tbd} methods use the public YOLOX detector, while \textit{e2e} methods employ their own detectors.}
\label{tab:results_test_dancetrack}
\end{table*}

\textbf{SportsMOT}. SportsMOT\footnote{SportsMOT GitHub page: \url{https://github.com/MCG-NJU/SportsMOT}.}
 is a large-scale MOT dataset spanning basketball, volleyball, and football. It is challenging due to visually similar players and highly non-linear motion with frequent accelerations, decelerations, and direction changes, as well as diverse match scenarios and camera viewpoints~\cite{sportsmot}. On this benchmark, tracking-by-detection methods combining heuristics with learnable motion models perform strongly~\cite{movesort,deepmovesort,ettrack,motiontrack}, often outperforming current end-to-end trackers.

As shown in Table~\ref{tab:results_test_sportsmot}, TDLP outperforms both end-to-end and tracking-by-detection methods on SportsMOT, establishing a new state-of-the-art result. Compared to the previous best method, CAMELTrack, TDLP improves HOTA by $1.5\%$ and IDF1 by $3.1\%$. The larger gains in association metrics relative to DanceTrack are expected, as SportsMOT places greater emphasis on accurate motion modeling due to the highly non-linear player trajectories. In this setting, TDLP’s stronger ability to model motion yields more pronounced benefits. Notably, the lightweight \textit{TDLP-bbox} variant outperforms all prior motion-only methods without relying on appearance features or validation-based detector tuning.

\begin{table*}
\scriptsize
\centering
\begin{tabular}{lcccccc}
Method & Training setup & HOTA & DetA & AssA & MOTA & IDF1 \\
\hline
\textit{e2e} & & \multicolumn{5}{c}{} \\
MeMOTR~\cite{memotr} & Train & 70.0 & 83.1 & 59.1 & 91.5 & 71.4 \\
MOTIP~\cite{motip} & Train & 71.9 & 83.4 & 62.0 & 92.9 & 75.0 \\
\hline
\textit{tbd (bbox features)} & & \multicolumn{5}{c}{} \\
ByteTrack~\cite{bytetrack} & Train & 62.8 & 77.1 & 51.2 & 94.1 & 69.8 \\
OC\_SORT~\cite{ocsort} & Train & 71.9 & 86.4 & 59.8 & 94.5 & 72.2 \\
MoveSORT~\cite{movesort} & Train+Val & 74.6 & 87.5 & 63.7 & 96.7 & 76.9 \\
\textbf{TDLP-bbox (ours)} & Train & 74.8 & 87.2 & 64.1 & 95.4 & 79.0 \\
\hline
\textit{tbd (extra features)} & & \multicolumn{5}{c}{} \\
Deep-EIoU~\cite{deep_eiou} & Train & 74.1 & 87.2 & 63.1 & 95.1 & 75.0 \\
Deep-EIoU~\cite{deep_eiou} & Train+Val & 77.2 & \textit{88.2} & 67.7 & \textit{96.3} & 79.8 \\
DeepMoveSORT~\cite{deepmovesort} & Train+Val& 78.7 & 88.1 & 70.3 & \textbf{96.5} & 81.7 \\
CAMELTrack~\cite{cameltrack} & Train & \textit{80.4} & \textbf{88.8} & \textit{72.8} & \textit{96.3} & \textit{84.8} \\
\textbf{TDLP (ours)} & Train & \textbf{81.9} & 88.0 & \textbf{76.3} & 95.8 & \textbf{87.5} \\
\end{tabular}
\caption{Evaluation results on the SportsMOT test set. The acronyms \textit{e2e} and \textit{tbd} denote end-to-end and tracking-by-detection methods. \textit{Training setup} indicates whether the validation set was additionally used during training for the submitted solutions~\cite{sportsmot}.}
\label{tab:results_test_sportsmot}
\end{table*}

\textbf{BEE24}. BEE24\footnote{BEE24 (TOPICTrack) GitHub page: \url{https://github.com/holmescao/TOPICTrack}.}
 is a recent MOT benchmark for ecological scenarios, comprising video sequences of bees recorded under natural conditions. It is highly challenging due to crowded scenes, frequent occlusions, rapid and irregular motion, and near-identical appearances. By introducing a non-human domain with fundamentally different visual and motion characteristics, BEE24 complements existing human-focused MOT benchmarks and provides a valuable testbed for evaluating robustness and generalization.

Table~\ref{tab:results_test_bee24} reports results on BEE24, where TDLP improves over the previous state-of-the-art method by $1.6\%$ HOTA and $3.0\%$ IDF1. The gains in association accuracy are particularly notable, as BEE24 is the most challenging benchmark in terms of motion prediction, with dense interactions, occlusions, and highly irregular trajectories. Unlike human-tracking datasets, we rely solely on the bounding box features, since keypoints and appearance embeddings are unavailable and appearance cues are inherently limited. While orientation could be informative~\cite{beehive_dataset}, it is not provided, further highlighting the effectiveness of TDLP’s motion modeling.

\begin{table*}
\scriptsize
\centering
\begin{tabular}{lcccc}
Method & HOTA & AssA & IDF1 & MOTA \\
\hline
\textit{e2e} & \multicolumn{4}{c}{} \\
TrackFormer~\cite{trackformer} & 44.3 & 42.3 & 53.9 & 41.5 \\
\hline
\textit{tbd (bbox features)} & \multicolumn{4}{c}{} \\
ByteTrack~\cite{bytetrack} & 43.2 & 38.3 & 56.8 & 59.2 \\
OC\_SORT~\cite{ocsort} & 42.7 & 36.8 & 55.3 & 61.6 \\
TOPICTrack~\cite{bee24} & 46.6 & 40.3 & 59.7 & 66.7 \\
CAMELTrack~\cite{cameltrack} & \textit{50.3} & \textit{42.6} & \textit{63.8} & \textbf{75.7} \\
\textbf{TDLP-bbox (ours)} & \textbf{51.9} & \textbf{46.2} & \textbf{66.8} & \textit{74.9} \\
\end{tabular}
\caption{Evaluation results on the BEE24 test set. All \textit{tbd} trackers use the same object detector.}
\label{tab:results_test_bee24}
\end{table*}

Additional benchmark results, including experiments on MOT17~\cite{mot_challenge} and SoccerNet~\cite{soccernet}, are provided in Appendix~\ref{appendix:additional_results}.

\subsection{Ablation Studies and Discussion}
\label{sec:ablation_studies}

\textbf{Component ablation study.} We use CTDP (contrastive track--detection learning) as the baseline, which shares the TDLP architecture but replaces the link prediction head with a contrastive objective. Implementation details are provided in Appendix~\ref{appendix:ctdp}. Following~\cite{cameltrack}, we adopt the InfoNCE loss~\cite{infonce} and train directly on clips without batch sampling, as batch sampling provides no measurable benefit. Results are shown in Table~\ref{tab:components_ablation_table}. We find that both the training objective and carefully designed motion features are critical for performance, with larger gains in the bounding box-only setting and smaller relative improvements when keypoints and appearance are included. Compared to CTDP, TDLP improves HOTA by $11.1\%$ (bbox-only) and $5.4\%$ (full modalities) on DanceTrack, and by $14.0\%$ and $11.3\%$, respectively, on SportsMOT.

\begin{table}[t]
\scriptsize
\centering
\begin{tabular}{lccccc}
\toprule
\multicolumn{2}{c}{Components} 
& \multicolumn{2}{c}{DanceTrack} 
& \multicolumn{2}{c}{SportsMOT} \\
\cmidrule(lr){1-2} \cmidrule(lr){3-4} \cmidrule(lr){5-6}
GFT & Method & Bb & MM & Bb & MM \\
\midrule
Yes & TDLP & \textbf{62.1} & \textbf{65.4} & \textbf{80.4} & \textbf{88.4} \\
Yes & CTDP & 56.7 & 62.3 & 71.8 & 76.8 \\
No  & CTDP & 51.0 & 60.1 & 66.4 & 77.1 \\
\bottomrule
\end{tabular}
\caption{Ablation of \textit{geometric feature transform} (GFT) and link prediction (TDLP).}
\label{tab:components_ablation_table}
\end{table}

\begin{table}[t]
\scriptsize
\centering
\begin{tabular}{cccccc}
\toprule
\multicolumn{3}{c}{Features} 
& \multicolumn{2}{c}{DanceTrack} 
& \multicolumn{1}{c}{SportsMOT} \\
\cmidrule(lr){1-3} \cmidrule(lr){4-5} \cmidrule(lr){6-6}
App & Kp & Bb & HOTA & IDF1 & HOTA / IDF1 \\
\midrule
\checkmark & \checkmark & \checkmark & \textbf{65.3} & \textbf{70.5} & \textbf{88.8 / 92.9} \\
\checkmark &            &            & 40.2 & 42.6 & 74.1 / 76.0 \\
           & \checkmark &            & 58.4 & 62.2 & 80.7 / 84.7 \\
           &            & \checkmark & 61.6 & 66.2 & 80.4 / 83.9 \\
\bottomrule
\end{tabular}
\caption{Feature-type ablation results for TDLP on DanceTrack and SportsMOT.}
\label{tab:feature_importance}
\end{table}

\textbf{Feature ablation study.} We analyze TDLP variants using individual feature modalities (Table~\ref{tab:feature_importance}). Combining all modalities yields the best performance. When used in isolation, bounding boxes and keypoints perform strongly, while appearance features alone are markedly weaker. This contrasts with~\cite{cameltrack}, where appearance cues are more effective, whereas our motion-based features contribute more substantially. We attribute the weaker appearance-only results to a mismatch between metric learning–based appearance pretraining and the link prediction training objective.

\begin{figure*}[t]
\centering
\subfigure[DanceTrack]{%
  \includegraphics[width=0.46\linewidth]{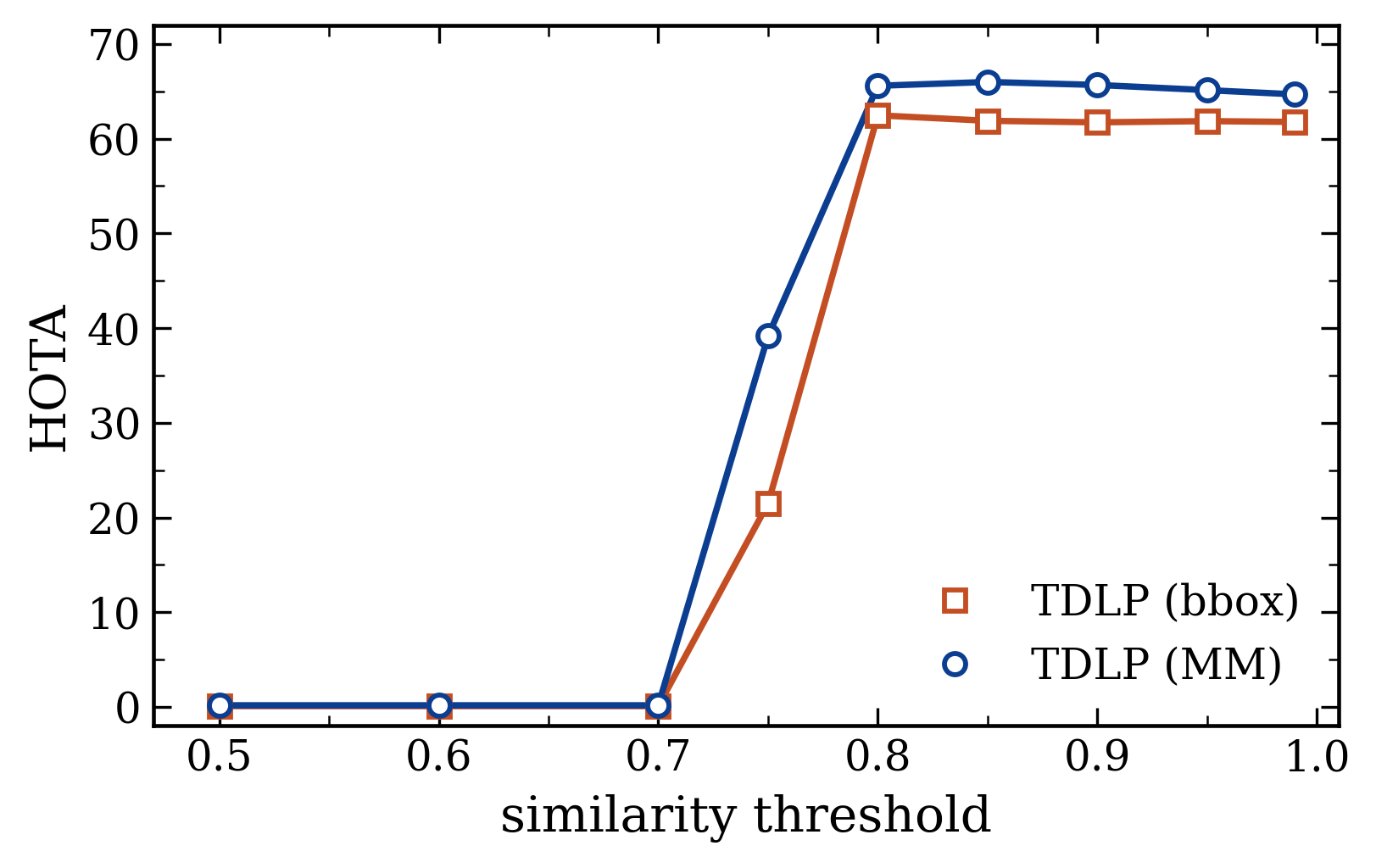}%
}
\hfill
\subfigure[SportsMOT]{%
  \includegraphics[width=0.46\linewidth]{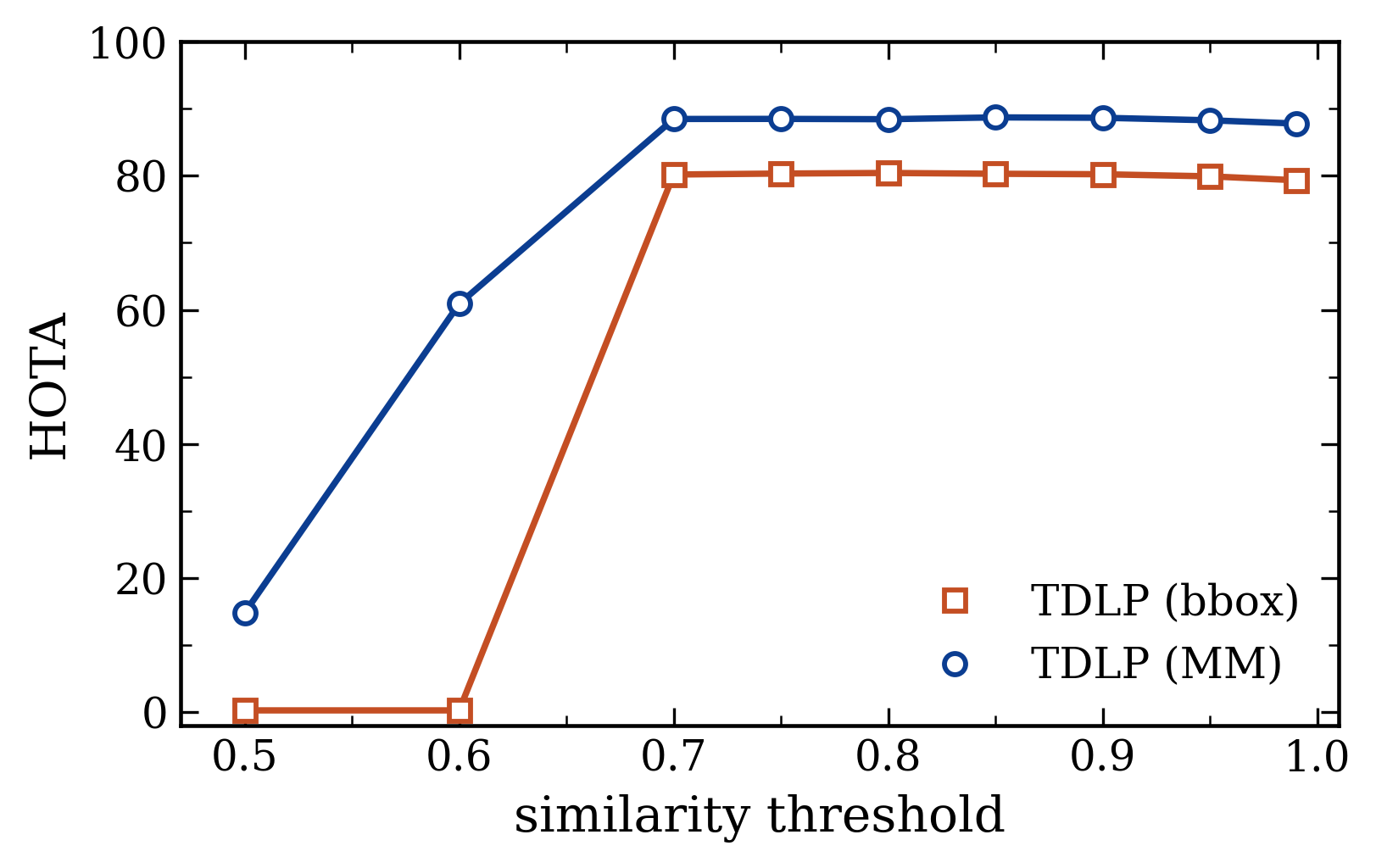}%
}
\caption{HOTA vs.\ similarity threshold $\theta_{\mathrm{sim}}$ on DanceTrack and SportsMOT validation. After a per-dataset cliff, HOTA stays within $\approx 1$ point across a wide plateau.}
\label{fig:sim_threshold_sensitivity}
\end{figure*}

\textbf{Similarity-threshold sensitivity.} To assess sensitivity to the similarity-threshold hyperparameter, we sweep the threshold $\theta_{\mathrm{sim}}$ on the DanceTrack and SportsMOT validation sets (Figure~\ref{fig:sim_threshold_sensitivity}). Both datasets show the same pattern: a sharp drop at low thresholds, followed by a broad stable plateau. The default value $\theta_{\mathrm{sim}}=0.9$ transfers well across both datasets, while a coarse sweep with a step size of $0.1$ is sufficient to identify the plateau on a new dataset. These results indicate that TDLP is not highly sensitive to this hyperparameter, supporting its ability to generalize across domains.

\begin{figure}[t]
\centering
\includegraphics[width=0.85\linewidth]{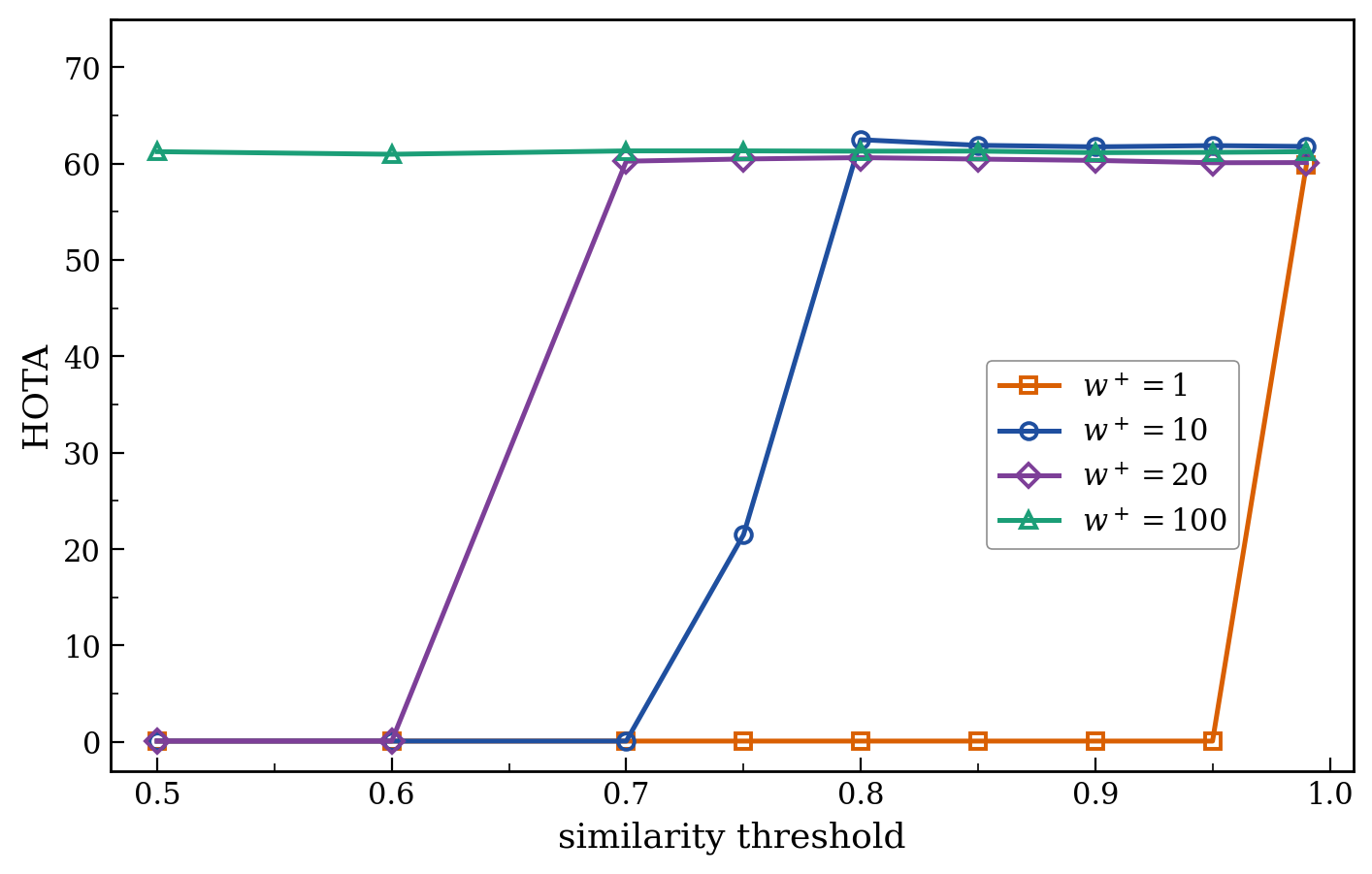}
\caption{HOTA vs.\ $\theta_{\mathrm{sim}}$ for $w^{+}\in\{1,10,20,100\}$ on DanceTrack validation (TDLP-bbox). Larger $w^{+}$ widens the plateau; $w^{+}=10$ has the highest peak.}
\label{fig:posweight_sensitivity}
\end{figure}

\textbf{Positive-class weight sensitivity.} The BCE loss uses a positive-class weight $w^{+}$ to counter the track--detection pair imbalance. We sweep $w^{+}\in\{1,10,20,100\}$ on the DanceTrack validation set for TDLP-bbox, and report its effect on similarity-threshold sensitivity in Figure~\ref{fig:posweight_sensitivity} and on overall performance in Table~\ref{tab:posweight_ablation}. The weight $w^{+}$ mainly controls the \emph{width} of the usable $\theta_{\mathrm{sim}}$ plateau rather than the peak performance: $w^{+}=1$ works well only at $\theta_{\mathrm{sim}}=0.99$, whereas $w^{+}=100$ remains stable down to $\theta_{\mathrm{sim}}=0.5$. We use $w^{+}=10$, as it achieves the highest peak performance while reducing sensitivity to the threshold configuration.

\begin{table}[t]
\scriptsize
\centering
\setlength{\tabcolsep}{4pt}
\begin{tabular}{cccc}
\toprule
$w^{+}$ & HOTA & $\theta_{\mathrm{sim}}$ & Plateau \\
\midrule
$1$   & $59.81$          & $0.99$          & $\{0.99\}$ \\
$10$  & $\mathbf{62.47}$ & $\mathbf{0.80}$ & $[0.80, 0.99]$ \\
$20$  & $60.61$          & $0.80$          & $[0.70, 0.99]$ \\
$100$ & $61.31$          & $0.70$          & $[0.50, 0.99]$ \\
\bottomrule
\end{tabular}
\caption{Effect of $w^{+}$ on DanceTrack validation (TDLP-bbox; identical trend on multi-modal TDLP). Larger $w^{+}$ widens the plateau at a small peak cost.}
\label{tab:posweight_ablation}
\end{table}

\textbf{Link prediction head ablation.} We compare four head designs on the DanceTrack validation set in the bbox-only setting (Table~\ref{tab:link_prediction_head_ablation}). The three MLP-based heads---the full $[z_t, z_d, |z_t - z_d|]$ design from Equation~10, a $|z_t - z_d|$-only variant, and a $[z_t, z_d]$ variant---perform within $0.4$ HOTA of one another, showing that any MLP head that mixes the embeddings non-linearly reaches a similar ceiling. In contrast, the dot product head, which reduces the pair to a single scalar $\tau(z_t \cdot z_d) + b$, loses approximately $5$ HOTA due to information loss before link prediction.

\begin{table}[t]
\scriptsize
\centering
\begin{tabular}{lcc}
\toprule
Head & Params & Peak HOTA \\
\midrule
$\mathrm{MLP}([z_t, z_d, |z_t - z_d|])$ (Eq.~\eqref{eq:tdlp_pairwise_representation}) & $1.5$M & $62.47$ \\
$\mathrm{MLP}(|z_t - z_d|)$                                                           & $0.5$M & $\mathbf{62.85}$ \\
$\mathrm{MLP}([z_t, z_d])$                                                            & $1.0$M & $62.83$ \\
$\tau (z_t \cdot z_d) + b$                                                            & $2$    & $57.39$ \\
\bottomrule
\end{tabular}
\caption{link prediction head variants on DanceTrack validation (bbox-only TDLP). MLP-based heads are interchangeable; the scalar dot-product head loses $\approx 5$ HOTA.}
\label{tab:link_prediction_head_ablation}
\end{table}

\textbf{Multi-modal fusion choice.} We replace the additive aggregator from Equation~\eqref{eq:tdlp_mm_fusion} with four alternatives on the DanceTrack validation set, while keeping the per-modality checkpoints frozen (Table~\ref{tab:mm_fusion_ablation}). All five variants perform within $0.72$ HOTA of one another, which is below the $\approx 1$ HOTA seed variance measured across multiple runs. This suggests that the fusion choice has little impact at the modeling level. We keep \texttt{Sum} due to its simplicity.

\begin{table}[t]
\scriptsize
\centering
\begin{tabular}{lcccc}
\toprule
Aggregator & Params & Peak HOTA & Peak $\theta_{\mathrm{sim}}$ & $\Delta$ \\
\midrule
\texttt{Sum} (Eq.~\eqref{eq:tdlp_mm_fusion}) & $0$    & $65.99$          & $0.85$ & $-$ \\
\texttt{LinearSum}                           & $3.1$M & $65.69$          & $0.80$ & $-0.30$ \\
\texttt{Concat+MLP}                          & $4.2$M & $\mathbf{66.11}$ & $0.85$ & $+0.12$ \\
\texttt{Attn} (gated)                        & $0.5$M & $65.84$          & $0.80$ & $-0.15$ \\
\texttt{Query} (cross-attn)                  & $3.1$M & $65.39$          & $0.80$ & $-0.60$ \\
\bottomrule
\end{tabular}
\caption{Fusion aggregators on DanceTrack validation (multi-modal TDLP). All variants are within $0.72$ HOTA, which is inside the seed-variance level. \texttt{Sum} is kept for its wider $\theta_{\mathrm{sim}}$ plateau, not for a higher peak.}
\label{tab:mm_fusion_ablation}
\end{table}

\begin{figure}
  \centering
  \includegraphics[width=0.70\linewidth]{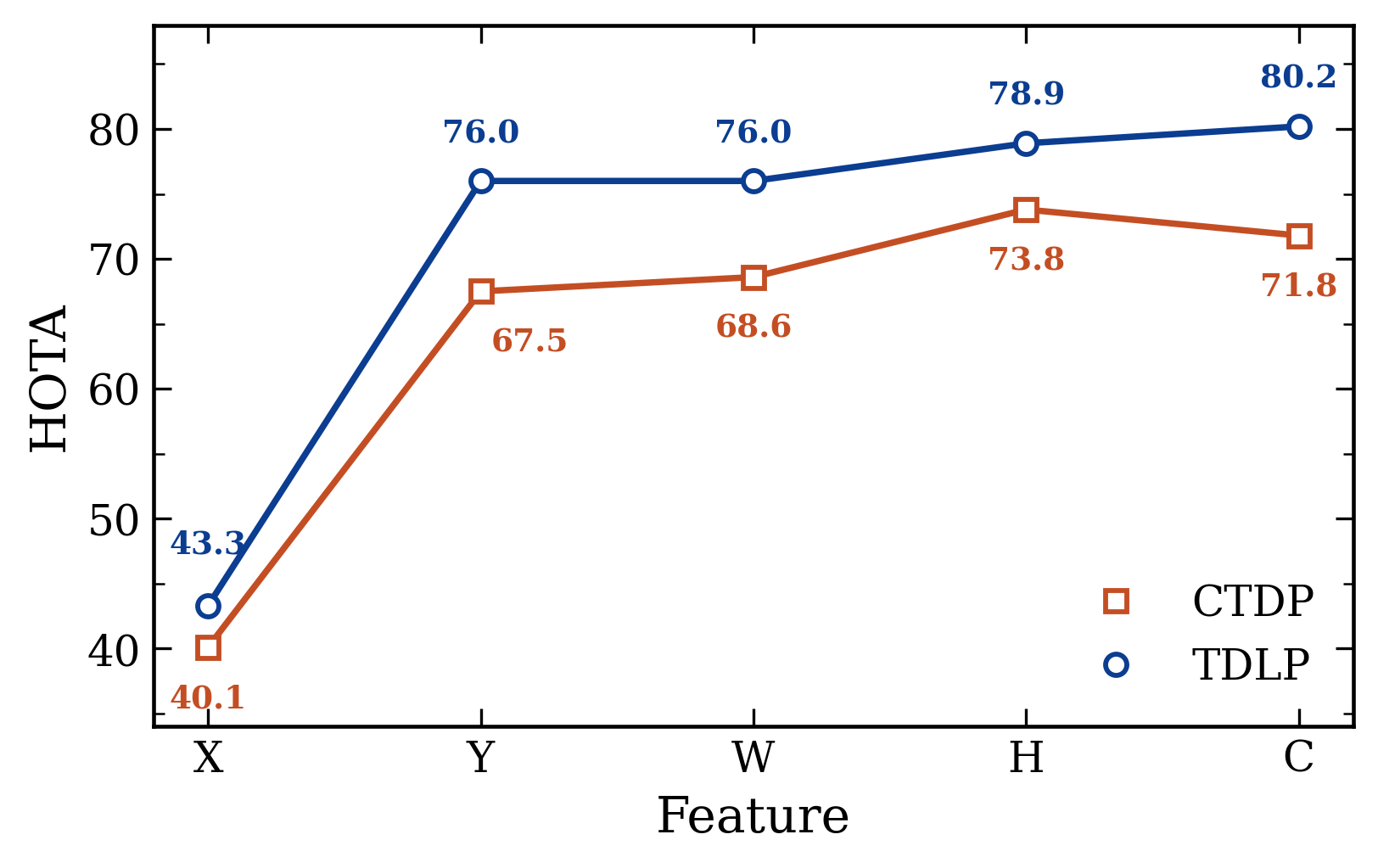}
  \caption{Comparison of CTDP and TDLP across different bounding box features (X, Y, W, H, C) measured by HOTA on the SportsMOT validation set.}
  \label{fig:bbox_features_ablation}
\end{figure}

\textbf{Object interaction encoder ablation.} TDLP uses two object interaction encoders $f_{\mathrm{inter}}$: a \emph{per-feature} encoder applied after each modality-specific temporal encoder, and an \emph{MM-level} encoder applied after multi-modal fusion. We perform a $2\times2$ ablation on DanceTrack validation by enabling these two encoders independently, with results reported in Table~\ref{tab:f_inter_ablation}. The per-feature encoder accounts for approximately $3\%$ HOTA, showing that object interaction before multi-modal fusion is important for stronger tracking performance. In contrast, the MM-level encoder contributes only $0.26\%$ HOTA when the per-feature encoder is present, which is within noise, and $1.22\%$ HOTA when the per-feature encoder is absent. These results indicate that the per-feature encoder is the load-bearing component, while the MM-level encoder mainly acts as a small redundant component and can be removed at inference without measurable loss.

\begin{table}[t]
\scriptsize
\centering
\begin{tabular}{lccc}
\toprule
Variant & Per-feat & MM-level & HOTA \\
\midrule
\texttt{Full}        & \checkmark & \checkmark & $\mathbf{65.99}$ \\
\texttt{PerFeatOnly} & \checkmark &            & $65.73$ \\
\texttt{MMOnly}      &            & \checkmark & $63.08$ \\
\texttt{None}        &            &            & $61.86$ \\
\bottomrule
\end{tabular}
\caption{Object interaction encoder ablation on DanceTrack validation. \texttt{Full} applies object interaction both before and after multi-modal fusion; \texttt{PerFeatOnly} applies it only before fusion; \texttt{MMOnly} applies it only after fusion; and \texttt{None} removes it. The pre-fusion encoder contributes $\approx 3$ HOTA, whereas the post-fusion encoder mainly provides a small redundancy buffer.}
\label{tab:f_inter_ablation}
\end{table}

\textbf{Analysis of link prediction compared to contrastive prediction.} We conduct additional experiments to understand why link prediction outperforms contrastive prediction under a fixed architecture. We focus on the bounding box–only setting using $(x, y, w, h, c)$, as these low-dimensional geometric features provide a controlled environment for isolating the learning objective. Starting from a single feature ($x$), we progressively add $y$, $w$, $h$, and $c$. As shown in Figure~\ref{fig:bbox_features_ablation}, TDLP consistently outperforms CTDP across all feature subsets, with the gap widening as more features are introduced. This highlights a fundamental difference between the objectives: contrastive learning enforces a global embedding structure in which all features must align within a shared metric space, whereas link prediction learns a local similarity function that can selectively weight informative components. As a result, TDLP benefits from richer feature combinations, while CTDP becomes increasingly constrained as heterogeneous features are added.

\begin{figure}[t]
\tiny
\centering
\setlength{\tabcolsep}{3pt}
\renewcommand{\arraystretch}{1.0}

\subfigure[CTDP rank test]{%
  \begin{tabular}{lcccc}
    motion pattern & $n_0$ & $n_1$ & $n_2$ & $n_3$ \\
    \midrule
    static              & . & . & . & . \\
    static conf decay & . & . & . & . \\
    linear              & . & . & . & . \\
    linear conf decay & . & . & . & . \\
    nonlinear accel    & . & . & . & . \\
    nonlinear curve    & . & X & . & X \\
    \bottomrule
  \end{tabular}
}
\hspace{0.01\textwidth}
\subfigure[TDLP rank test]{%
  \begin{tabular}{lcccc}
    motion pattern & $n_0$ & $n_1$ & $n_2$ & $n_3$ \\
    \midrule
    static              & . & . & . & . \\
    static conf decay & . & . & . & . \\
    linear              & . & . & . & . \\
    linear conf decay & . & . & . & . \\
    nonlinear accel    & . & . & . & . \\
    nonlinear curve    & . & . & . & . \\
    \bottomrule
  \end{tabular}
}

\subfigure[CTDP threshold test]{%
  \begin{tabular}{lcccc}
    motion pattern & $n_0$ & $n_1$ & $n_2$ & $n_3$ \\
    \midrule
    static              & X & X & X & X \\
    static conf decay & X & X & X & X \\
    linear              & X & X & X & X \\
    linear conf decay & X & X & X & X \\
    nonlinear accel    & X & X & X & X \\
    nonlinear curve    & X & X & X & X \\
    \bottomrule
  \end{tabular}
}
\hspace{0.01\textwidth}
\subfigure[TDLP threshold test]{%
  \begin{tabular}{lcccc}
    motion pattern & $n_0$ & $n_1$ & $n_2$ & $n_3$ \\
    \midrule
    static              & . & . & . & . \\
    static conf decay & . & . & . & . \\
    linear              & . & . & . & . \\
    linear conf decay & . & . & . & . \\
    nonlinear accel    & . & . & . & . \\
    nonlinear curve    & . & . & X & . \\
    \bottomrule
  \end{tabular}
}

\caption{Pass/fail matrices for CTDP and TDLP. Dots indicate correct outcomes; \texttt{X} marks failures leading to ID switches. See Appendix~\ref{appendix:additional_results}.}
\label{fig:combined_matrices}
\end{figure}

To further analyze why TDLP outperforms CTDP in the single-feature setting (Figure~\ref{fig:bbox_features_ablation}), we design a controlled experiment that isolates association behavior. A single track is observed over a short history and matched to one positive detection (its true continuation) and several systematically varied negative detections. We evaluate two failure modes during linear assignment: (i) a \emph{rank test}, which checks whether the positive detection is ranked above all negatives, and (ii) a \emph{threshold test}, which verifies whether negatives remain below the association threshold when the positive detection is absent. Failing the rank test guarantees an ID switch, as a negative detection is preferred over the true continuation. Failing the threshold test leads to an ID switch when the true detection is missing, since negatives are not rejected by the gating mechanism during linear assignment.

Figure~\ref{fig:combined_matrices} highlights a clear contrast between CTDP and TDLP. Both methods perform well in rank tests, indicating similar behavior when detector recall is high. However, only TDLP remains reliable in threshold tests: CTDP fails to reject negative detections across all configurations, revealing strong sensitivity to detector false negatives. In contrast, TDLP consistently suppresses negatives, demonstrating robustness in scenarios where linear assignment most often fails. This difference arises from the contrastive objective of CTDP, which maps spatially nearby boxes to similar embeddings, whereas TDLP maintains sharper discriminative boundaries. Full experimental details are provided in Appendix~\ref{appendix:additional_results}.

\section{Conclusion}
\label{sec:conclusion}

We propose Track–Detection link prediction (TDLP), a learning-based data association framework that formulates multi-object tracking as a per-frame link prediction between tracks and detections. TDLP learns associations directly from data, avoids handcrafted heuristics, and remains more computationally efficient than fully end-to-end trackers while preserving modularity of tracking-by-detection pipelines. Extensive experiments on multiple challenging benchmarks show that TDLP outperforms state-of-the-art heuristic-based and end-to-end approaches, particularly in scenarios with non-linear motion. Even when using only bounding box features, TDLP achieves strong performance, highlighting the effectiveness of link prediction for geometric association. Through controlled analysis, we expose the limitations of metric learning–based association. TDLP’s main limitation is its computational cost, which arises from transformer-based encoders and feature extractors in case any additional feature modalities are exploited. Future work will explore longer temporal windows and architectural optimizations to reduce this cost.

\appendix

\section{Implementation details}
\label{appendix:implementation}

\textbf{Architecture.} Figure~\ref{fig:architecture} illustrates the TDLP architecture. For bounding boxes and pose keypoints, both static and motion encoders are implemented as MLPs with two linear layers, LayerNorm~\cite{layernorm}, SiLU~\cite{silu}, and dropout~\cite{dropout}. Appearance features use only a static encoder following CAMELTrack~\cite{cameltrack}. All feature embeddings are $512$-dimensional and are processed by modality-specific transformer-based temporal encoders~\cite{object_tracking_in_robotics} with $4$ layers and $8$ heads. The resulting features are projected to $1024$ dimensions, summed into a multi-modal representation, and passed through an object interaction encoder with the same architecture. The link prediction head is a two-layer MLP with LayerNorm and SiLU.

\textbf{Training.} For each feature modality, TDLP is pretrained using AdamW~\cite{adam,adamw} with a cosine annealing scheduler~\cite{cosan}, $2$ warm-up epochs, a learning rate $5{\times}10^{-2}$, a weight decay $1{\times}10^{-2}$, and gradient clipping at $1.0$. The multi-modal model is then fine-tuned for $10$ epochs with $1$ warm-up epoch, a learning rate $1{\times}10^{-5}$, and a weight decay $1{\times}10^{-3}$. The BCE positive class weight is set to $10$. The clip length is $50$ for DanceTrack, MOTChallenge, and BEE24, and $150$ for SportsMOT and SoccerNet to support long-term occlusions, making TDLP the first online tracker to handle this setting.

\textbf{Object detection and feature extraction models.} We use the official YOLOX detector for DanceTrack~\cite{yolox}, the dataset-provided YOLOX detector for SportsMOT and SoccerNet~\cite{sportsmot}, TOPICTrack’s detector for BEE24~\cite{bee24}, and ByteTrack’s YOLOX detector for all MOTChallenge benchmarks. To ensure a transparent comparison, TDLP uses the same pose keypoints and appearance features as CAMELTrack~\cite{cameltrack}, differing only in the association mechanism.

\textbf{Tracker hyper-parameters.} The definitions of these hyperparameters are provided in Section~\ref{sec:tdlp_inference}. 
Values are listed in the following order: DanceTrack, SportsMOT, BEE24, and MOTChallenge, with SoccerNet matching SportsMOT. 
The detection threshold $\theta_{\mathrm{det}}$ is $0.4$, $0.1$, $0.6$, and $0.5$; the link threshold $\theta_{\mathrm{link}}$ is $0.015$, $0.01$, $0.65$, and $0.05$; the initialization time $T_{\mathrm{init}}$ is $3$, $1$, $0$, and $1$; the initialization confidence $\theta_{\mathrm{new}}$ is $0.9$, $0.4$, $0.6$, and $0.55$; and the maximum lost duration $T_{\mathrm{lost}}$ is $50$, $150$, $50$, and $50$ frames.

\section{CTDP baseline}
\label{appendix:ctdp}

CTDP (\emph{contrastive track--detection prediction}) is the metric learning baseline used in Section~\ref{sec:ablation_studies} to isolate the effect of the prediction formulation. It keeps the TDLP architecture of Section~\ref{sec:architecture} unchanged up to the refined track and detection embeddings $\bar{\bm{z}}^{\mathrm{trk}}_i$ and $\bar{\bm{z}}^{\mathrm{det}}_j$: the geometric feature transform, static/motion encoders, temporal transformer, multi-modal fusion, and object interaction encoder $f_{\mathrm{inter}}$ all use the same hyper-parameters as TDLP. Only the scoring head and training loss are changed.

Instead of forming the pairwise representation of Eq.~\eqref{eq:tdlp_pairwise_representation} and applying the link prediction MLP $\phi$, CTDP projects the track embedding to the detection embedding space, $L_2$-normalizes both embeddings, and scores each pair by scaled cosine similarity:
\begin{equation}
    \tilde{\bm{z}}^{\mathrm{trk}}_i =
    \frac{\mathbf{P}\bar{\bm{z}}^{\mathrm{trk}}_i}
         {\lVert \mathbf{P}\bar{\bm{z}}^{\mathrm{trk}}_i \rVert_2},
    \qquad
    \tilde{\bm{z}}^{\mathrm{det}}_j =
    \frac{\bar{\bm{z}}^{\mathrm{det}}_j}
         {\lVert \bar{\bm{z}}^{\mathrm{det}}_j \rVert_2},
    \qquad
    S_{ij} = \tfrac{1}{2}\!\left(1 +
    \langle \tilde{\bm{z}}^{\mathrm{trk}}_i,
            \tilde{\bm{z}}^{\mathrm{det}}_j \rangle \right).
    \label{eq:ctdp_score}
\end{equation}
The score $S_{ij}\in[0,1]$ replaces the TDLP link probability and is used by the same Hungarian-matching inference pipeline.

CTDP is trained with clip-level InfoNCE~\cite{infonce} over the normalized embeddings:
\begin{equation}
    \mathcal{L}_{\mathrm{NCE}}
    = - \sum_{i=1}^{N_t}
      \log \frac{
      \exp\!\left(
      \langle \tilde{\bm{z}}^{\mathrm{trk}}_i,
              \tilde{\bm{z}}^{\mathrm{det}}_{\pi(i)} \rangle / \tau
      \right)}
      {\sum_{k \in \mathcal{N}(i)}
      \exp\!\left(
      \langle \tilde{\bm{z}}^{\mathrm{trk}}_i,
              \tilde{\bm{z}}^{\mathrm{det}}_k \rangle / \tau
      \right)},
    \label{eq:ctdp_nce_loss}
\end{equation}
where $\pi(i)$ is the positive continuation detection for track $i$, $\mathcal{N}(i)$ contains the within-clip negatives, and $\tau=0.07$ is the temperature. Training uses sampled clips directly, without a memory bank, queue, momentum encoder, or additional batch-level negative sampling.

All optimization settings are identical to TDLP, including the optimizer, learning rate, weight decay, scheduler, gradient clipping, clip length, and batch size. Since CTDP and TDLP share the full forward pass up to $f_{\mathrm{inter}}$, and their final heads contribute less than $0.5\%$ of the forward latency in our setting, performance differences mainly reflect the prediction formulation rather than architectural or runtime differences.

\section{Additional results}
\label{appendix:additional_results}

\textbf{Additional benchmarks.}
The MOTChallenge~\cite{mot_challenge} benchmark consists of short video sequences with mostly linear pedestrian motion and limited appearance variation, and lacks an official validation split.
We report results on MOT17 in Table~\ref{tab:mot17}.
On this dataset, our model underperforms compared to heuristic-based methods, primarily due to the limited amount of training data and the absence of a validation split, which prevents proper hyper-parameter tuning. Heuristic-based trackers commonly split sequences into two halves~\cite{bytetrack,botsort,ocsort,deepocsort,movesort,deepmovesort} to mitigate these issues; however, this strategy is suboptimal in our case, as the model could overfit to appearance cues in later segments of the sequence.
Metric learning-based approaches rely on inter-clip sampling, which partially alleviates these limitations, as reflected in Table~\ref{tab:mot17}, where CAMELTrack outperforms our method by $1.8\%$ in terms of HOTA. We leave a more thorough investigation of these limitations to future work.

For SoccerNet~\cite{soccernet}, the dataset comprises video clips of soccer matches with the goal of tracking players, making it closely related to SportsMOT in terms of domain and visual characteristics. The results are reported in Table~\ref{tab:soccernet}. TDLP achieves the best overall performance across all metrics, while TDLP-bbox consistently outperforms all methods that rely solely on detector bounding box features. In particular, TDLP-bbox improves HOTA by $1.9\%$ over the strongest bbox-only baseline and TDLP outperforms CAMELTrack by $2.1\%$ in terms of HOTA.

\begin{table}[t]
\scriptsize
\centering
\begin{tabular}{lccccc}
\toprule
Method & HOTA & DetA & AssA & MOTA & IDF1 \\
\midrule
\multicolumn{6}{l}{\textit{e2e}} \\
MOTR~\cite{motr}     & 57.8 & 60.3 & 55.7 & 73.4 & 68.6 \\
MeMOTR~\cite{memotr} & 58.8 & 59.6 & 58.4 & 72.8 & 71.5 \\
MOTIP~\cite{motip}   & 59.2 & 62.0 & 56.9 & 75.5 & 71.2 \\
\midrule
\multicolumn{6}{l}{\textit{tbd}} \\
FairMOT~\cite{fairmot}       & 59.3 & -- & -- & 73.7 & 72.3 \\
OC-SORT~\cite{ocsort}        & 61.7 & -- & -- & 76.0 & 76.2 \\
ByteTrack~\cite{bytetrack}   & \textbf{62.8} & -- & -- & \textbf{78.7} & \textbf{77.1} \\
GHOST~\cite{ghost}           & \textbf{62.8} & -- & -- & \textbf{78.7} & \textbf{77.1} \\
CAMELTrack~\cite{cameltrack} & \textit{62.4} & \textit{63.6} & \textbf{61.4} & 78.5 & \textit{76.5} \\
\textbf{TDLP (ours)}         & 60.6 & 63.4 & 58.2 & 78.0 & 73.7 \\
\bottomrule
\end{tabular}
\caption{Evaluation results on the MOT17 test set.}
\label{tab:mot17}
\end{table}

\begin{table}[t]
\scriptsize
\centering
\begin{tabular}{lccccc}
\toprule
Method & HOTA & DetA & AssA & MOTA & IDF1 \\
\midrule
\multicolumn{6}{l}{\textit{tbd (bbox features)}} \\
SORT~\cite{sort}             & 48.3 & \textbf{58.9} & 39.7 & \textit{68.6} & 53.4 \\
ByteTrack~\cite{bytetrack}   & 50.3 & \textbf{58.9} & 43.1 & \textbf{68.8} & 57.2 \\
\textbf{TDLP-bbox (ours)}    & 52.2 & 58.5 & 46.7 & 67.7 & 63.6 \\
\midrule
\multicolumn{6}{l}{\textit{tbd (extra features)}} \\
CAMELTrack~\cite{cameltrack} & 54.2 & \textit{58.7} & 50.3 & 67.4 & \textit{67.5} \\
\textbf{TDLP (ours)}         & \textbf{56.3} & \textit{58.7} & \textbf{54.2} & 67.4 & \textbf{70.4} \\
\bottomrule
\end{tabular}
\caption{Evaluation results on the SoccerNet test set.}
\label{tab:soccernet}
\end{table}

\begin{figure*}[t]
\centering
\includegraphics[width=0.98\linewidth]{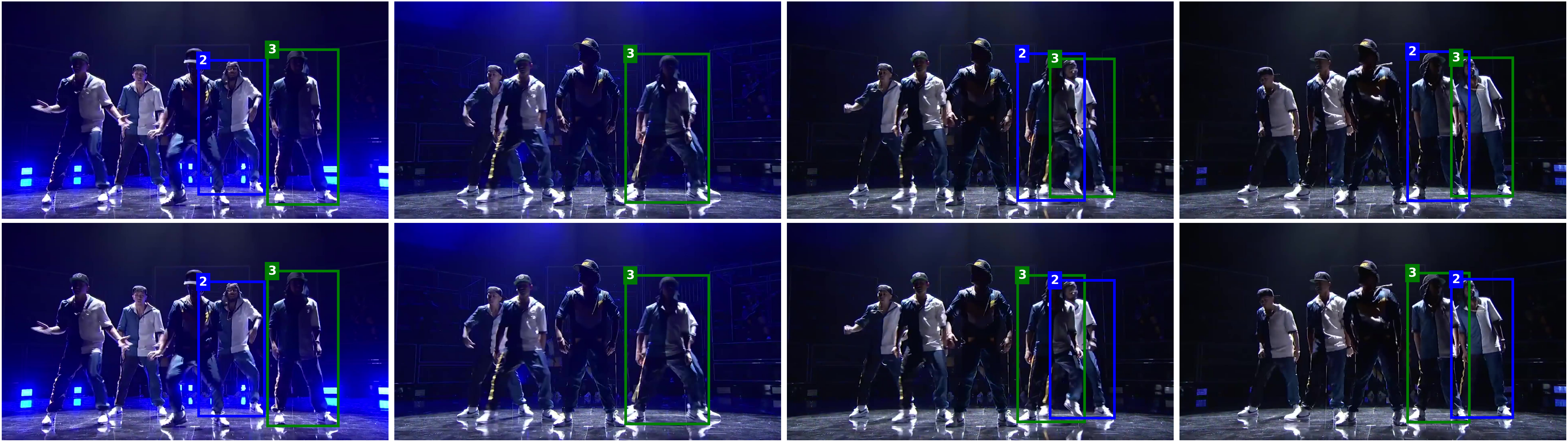}
\caption{\textbf{ID switch on the DanceTrack validation sequence \texttt{dancetrack0018}} for frames $\{360, 374, 376, 390\}$. ByteTrack is shown on top and TDLP-bbox on the bottom. Two similarly dressed dancers cross shoulder-to-shoulder between frames $374$ and $376$. ByteTrack swaps their identities after the crossing: ID~$3$ (green) reappears as ID~$2$ (blue), while ID~$2$ takes over ID~$3$. Because the post-crossing detections have nearly symmetric geometric overlap with the previous tracks, ByteTrack's IoU/Kalman-based association cannot disambiguate them. TDLP-bbox preserves both identities by conditioning association on motion-context features rather than only instantaneous geometric overlap.}
\label{fig:qualitative_dancetrack0018}
\end{figure*}

\begin{figure*}[t]
\centering
\includegraphics[width=0.98\linewidth]{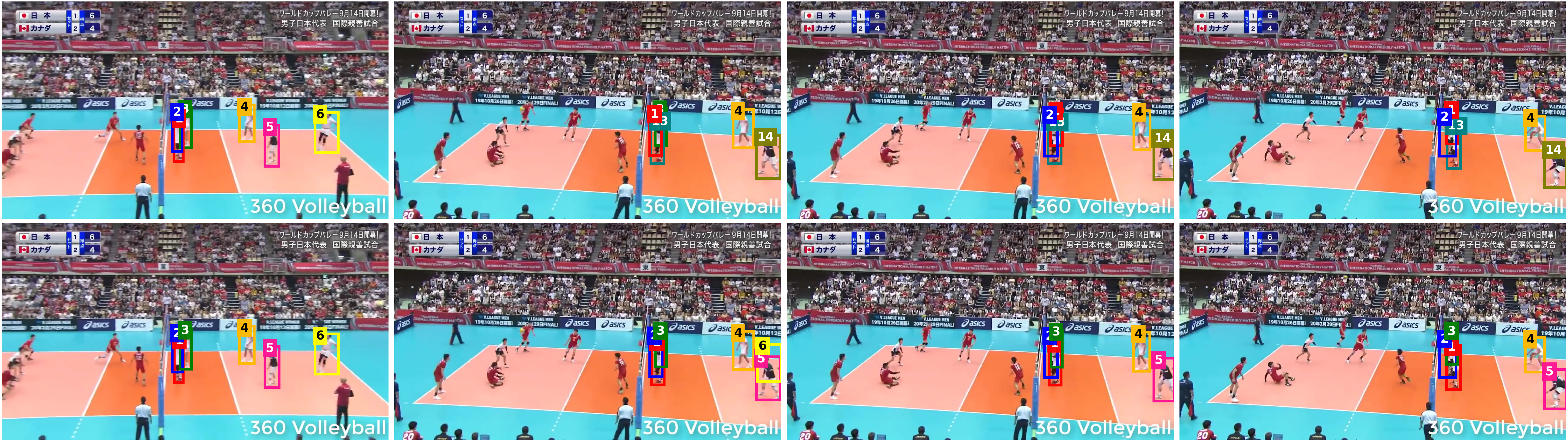}
\caption{\textbf{Track rebinds on SportsMOT-val \texttt{v\_0kUtTtmLaJA\_c007}} for frames $\{44, 58, 60, 74\}$. ByteTrack is shown on top and TDLP-bbox on the bottom. In this volleyball rally, visually similar players frequently occlude one another. TDLP-bbox maintains a consistent six-player labeling $\{1,2,3,4,5,6\}$ across the sampled frames. ByteTrack drops several players during short occlusions and later reacquires them as new track IDs, visible as the extra labels $13$ and $14$ in the top row.}
\label{fig:qualitative_volleyball}
\end{figure*}

\textbf{Qualitative comparison with ByteTrack}. Figures~\ref{fig:qualitative_dancetrack0018} and~\ref{fig:qualitative_volleyball} show two representative validation cases in which TDLP-bbox preserves identities that ByteTrack loses. Each figure contains a $2\times4$ panel from a single validation sequence, ordered left-to-right in time. The top row shows ByteTrack predictions, and the bottom row shows TDLP-bbox predictions. Both trackers use the same ByteTrack-compatible YOLOX detections. To reduce clutter and emphasize the illustrated failure mode, we omit predicted boxes that both trackers follow correctly and stably across all four frames. Thus, only tracks involved in the event of interest are shown. The omitted boxes are cases where the two trackers agree and would appear identically in both rows.
\medskip

\textbf{Computational efficiency.}
We evaluate the computational efficiency of TDLP-bbox and TDLP on DanceTrack and SportsMOT using a consumer-grade GPU. The results are shown in Table~\ref{tab:runtime_breakdown_dancetrack} for the DanceTrack dataset and Table~\ref{tab:runtime_breakdown_sportsmot} for the SportsMOT dataset. For reference, we also report CAMELTrack under the same setup, as it is up to twice as fast as end-to-end methods~\cite{cameltrack,motr}. Our profiling shows that the main computational bottleneck in TDLP is the temporal transformer, followed by the object interaction encoder. This bottleneck explains the drop in TDLP’s FPS when moving from DanceTrack to SportsMOT, since we use a longer history on SportsMOT: 150 frames compared with 60 frames on DanceTrack. Specifically, the prediction cost scales quadratically with the temporal history length and the number of objects, as in CAMELTrack. TDLP-bbox achieves substantially faster association than both the multi-modal TDLP and CAMELTrack variants.

\begin{table}[t]
\scriptsize
\centering
\begin{tabular}{lccccc}
\toprule
Method & Det & KP & App & Assoc & E2E \\
\midrule
TDLP-bbox (ours) 
& 9.9 & --  & --  & \textbf{240.7} & \textbf{9.5} \\
TDLP (ours)      
& 9.9 & 12.5 & 43.1 & 34.5 & 4.3 \\
CAMELTrack~\cite{cameltrack} 
& 9.9 & 12.5 & 43.1 & 32.2 & 4.3 \\
\bottomrule
\end{tabular}
\caption{Computational efficiency comparison on DanceTrack in terms of FPS. \textit{Det}, \textit{KP}, and \textit{App} denote detection, keypoint, and appearance extraction, while \textit{Assoc} and \textit{E2E} report association-only and end-to-end inference speed.}
\label{tab:runtime_breakdown_dancetrack}
\end{table}

\begin{table}[t]
\scriptsize
\centering
\begin{tabular}{lccccc}
\toprule
Method & Det & KP & App & Assoc & E2E \\
\midrule
TDLP-bbox (ours) 
& 9.1 & --  & --  & \textbf{60.4} & \textbf{7.9} \\
TDLP (ours)      
& 9.1 & 3.6  & 39.3 & 11.4 & 2.0 \\
CAMELTrack~\cite{cameltrack} 
& 9.1 & 3.6  & 39.3 & 20.1 & 2.2 \\
\bottomrule
\end{tabular}
\caption{Computational efficiency comparison on SportsMOT in terms of FPS. \textit{Det}, \textit{KP}, and \textit{App} denote detection, keypoint, and appearance extraction, while \textit{Assoc} and \textit{E2E} report association-only and end-to-end inference speed.}
\label{tab:runtime_breakdown_sportsmot}
\end{table}

\textbf{CTDP vs. TDLP experiment settings.} We use a 50-frame track history with base box $(0.30, 0.20, 0.05, 0.10, 0.95)$. Negative detections are generated using offsets $(s,s,s,s,0)$, $(s,0,0,0,0)$, $(0,s,0,0,0)$, and $(s/2,s/2,0,0,-5s)$ with $s=0.1$. Motion patterns include static, static with confidence decay ($\Delta c=-0.02$), linear $(0.01,0.008)$, linear horizontal drift ($\Delta x=0.003$, $\Delta c=-0.01$), nonlinear acceleration ($a_x=0.0001$, $a_y=0.00005$), and curved motion ($\Delta x=0.002$, $a_y=-0.0003$).

\bibliographystyle{elsarticle-num}
\bibliography{main}

@InProceedings{bytetrack,
  author    = {Yifu Zhang and Peize Sun and Yi Jiang and Dongdong Yu and Fucheng Weng and Zehuan Yuan and Ping Luo and Wenyu Liu and Xinggang Wang},
  editor    = {Shai Avidan and Gabriel Brostow and Moustapha Ciss{\'e} and Giovanni Maria Farinella and Tal Hassner},
  title     = {ByteTrack: Multi-object Tracking by Associating Every Detection Box},
  booktitle = {Computer Vision -- ECCV 2022},
  year      = {2022},
  publisher = {Springer Nature Switzerland},
  address   = {Cham},
  pages     = {1--21},
  doi       = {10.1007/978-3-031-20047-2_1},
  isbn      = {978-3-031-20047-2}
}

@INPROCEEDINGS{deepocsort,
  author={Maggiolino, Gerard and Ahmad, Adnan and Cao, Jinkun and Kitani, Kris},
  booktitle={2023 IEEE International Conference on Image Processing (ICIP)}, 
  title={Deep OC-Sort: Multi-Pedestrian Tracking by Adaptive Re-Identification}, 
  year={2023},
  pages={3025-3029},
  doi={10.1109/ICIP49359.2023.10222576}
}

@INPROCEEDINGS{ocsort,
  author={Cao, Jinkun and Pang, Jiangmiao and Weng, Xinshuo and Khirodkar, Rawal and Kitani, Kris},
  booktitle={2023 IEEE/CVF Conference on Computer Vision and Pattern Recognition (CVPR)}, 
  title={Observation-Centric SORT: Rethinking SORT for Robust Multi-Object Tracking}, 
  year={2023},
  pages={9686-9696},
  doi={10.1109/CVPR52729.2023.00934}
}

@inproceedings{hybridsort,
  title={Hybrid-sort: Weak cues matter for online multi-object tracking},
  author={Yang, Mingzhan and Han, Guangxin and Yan, Bin and Zhang, Wenhua and Qi, Jinqing and Lu, Huchuan and Wang, Dong},
  booktitle={Proceedings of the AAAI Conference on Artificial Intelligence},
  volume={38},
  number={7},
  pages={6504--6512},
  year={2024},
  doi={10.1609/aaai.v38i7.28471}
}

@misc{cameltrack,
      title={CAMELTrack: Context-Aware Multi-cue ExpLoitation for Online Multi-Object Tracking}, 
      author={Vladimir Somers and Baptiste Standaert and Victor Joos and Alexandre Alahi and Christophe De Vleeschouwer},
      year={2025},
      eprint={2505.01257},
      archivePrefix={arXiv},
      primaryClass={cs.CV},
      url={https://arxiv.org/abs/2505.01257}, 
}

@article{deepmovesort,
  title={Engineering an Efficient Object Tracker for Non-Linear Motion},
  author={Adžemović, Momir and Tadić, Petar and Petrović, Aleksandar and Nikolić, Milan},
  journal={arXiv preprint arXiv:2407.00738},
  year={2024},
  url={https://arxiv.org/abs/2407.00738}
}

@misc{motip,
      title={Multiple Object Tracking as ID Prediction}, 
      author={Ruopeng Gao and Yijun Zhang and Limin Wang},
      year={2024},
      eprint={2403.16848},
      archivePrefix={arXiv},
      primaryClass={cs.CV},
      url={https://arxiv.org/abs/2403.16848}, 
}

@inproceedings{motr,
  title={MOTR: End-to-End Multiple-Object Tracking with TRansformer},
  author={Zeng, Fangao and Dong, Bin and Zhang, Yuang and Wang, Tiancai and Zhang, Xiangyu and Wei, Yichen},
  booktitle={European Conference on Computer Vision (ECCV)},
  year={2022}
}

@INPROCEEDINGS{memotr,
  author={Gao, Ruopeng and Wang, Limin},
  booktitle={2023 IEEE/CVF International Conference on Computer Vision (ICCV)}, 
  title={MeMOTR: Long-Term Memory-Augmented Transformer for Multi-Object Tracking}, 
  year={2023},
  volume={},
  number={},
  pages={9867-9876},
  doi={10.1109/ICCV51070.2023.00908}
}

@INPROCEEDINGS{deep_eiou,
    author = { Huang, Hsiang-Wei and Yang, Cheng-Yen and Sun, Jiacheng and Kim, Pyong-Kun and Kim, Kwang-Ju and Lee, Kyoungoh and Huang, Chung-I and Hwang, Jenq-Neng },
    booktitle = { 2024 IEEE/CVF Winter Conference on Applications of Computer Vision Workshops (WACVW) },
    title = {{ Iterative Scale-Up ExpansionIoU and Deep Features Association for Multi-Object Tracking in Sports }},
    year = {2024},
    pages = {163-172},
    doi = {10.1109/WACVW60836.2024.00024}
}

@INPROCEEDINGS{dancetrack,
  author={Sun, Peize and Cao, Jinkun and Jiang, Yi and Yuan, Zehuan and Bai, Song and Kitani, Kris and Luo, Ping},
  booktitle={2022 IEEE/CVF Conference on Computer Vision and Pattern Recognition (CVPR)}, 
  title={DanceTrack: Multi-Object Tracking in Uniform Appearance and Diverse Motion}, 
  year={2022},
  pages={20961-20970},
  doi={10.1109/CVPR52688.2022.02032}
}

@inproceedings{sportsmot,
  author={Cui, Yutao and Zeng, Chenkai and Zhao, Xiaoyu and Yang, Yichun and Wu, Gangshan and Wang, Limin},
  booktitle={2023 IEEE/CVF International Conference on Computer Vision (ICCV)}, 
  title={SportsMOT: A Large Multi-Object Tracking Dataset in Multiple Sports Scenes}, 
  year={2023},
  pages={9887-9897},
  doi={10.1109/ICCV51070.2023.00910}
}

@inproceedings{soccernet,
  title={SoccerNet-Tracking: Multiple Object Tracking Dataset and Benchmark in Soccer Videos},
  author={Cioppa, Anthony and Giancola, Silvio and Deliege, Adrien and Kang, Le and Zhou, Xin and Cheng, Zhiyu and Ghanem, Bernard and Van Droogenbroeck, Marc},
  booktitle={Proceedings of the IEEE/CVF Conference on Computer Vision and Pattern Recognition},
  pages={3491--3502},
  year={2022}
}

@article{mot_challenge,
  author    = {Patrick Dendorfer and Aljo{\v{s}}a O{\v{s}}ep and Anton Milan and Konrad Schindler and Daniel Cremers and Ian Reid and Stefan Roth and Laura Leal-Taix{\'e}},
  title     = {MOTChallenge: A Benchmark for Single-Camera Multiple Target Tracking},
  journal   = {International Journal of Computer Vision},
  volume    = {129},
  number    = {4},
  pages     = {845--881},
  year      = {2021},
  doi       = {10.1007/s11263-020-01393-0},
  url       = {https://doi.org/10.1007/s11263-020-01393-0},
  issn      = {1573-1405}
}

@ARTICLE{bee24,
  author={Cao, Xiaoyan and Zheng, Yiyao and Yao, Yao and Qin, Huapeng and Cao, Xiaoyu and Guo, Shihui},
  journal={IEEE Transactions on Image Processing},
  title={TOPIC: A Parallel Association Paradigm for Multi-Object Tracking Under Complex Motions and Diverse Scenes},
  year={2025},
  volume={34},
  pages={743-758},
  doi={10.1109/TIP.2025.3526066}
}

@INPROCEEDINGS{trackformer,
  author={Meinhardt, Tim and Kirillov, Alexander and Leal-Taixé, Laura and Feichtenhofer, Christoph},
  booktitle={2022 IEEE/CVF Conference on Computer Vision and Pattern Recognition (CVPR)}, 
  title={TrackFormer: Multi-Object Tracking with Transformers}, 
  year={2022},
  volume={},
  number={},
  pages={8834-8844},
  doi={10.1109/CVPR52688.2022.00864}
}

@inproceedings{botsort,
  title={BoT-SORT: Robust Associations Multi-Pedestrian Tracking},
  author={Nir Aharon and Roy Orfaig and Ben-Zion Bobrovsky},
  year={2022},
  booktitle={arXiv preprint},
  note = {Arxiv: 2206.14651}
}

@misc{yolox,
      title={YOLOX: Exceeding YOLO Series in 2021}, 
      author={Zheng Ge and Songtao Liu and Feng Wang and Zeming Li and Jian Sun},
      year={2021},
      eprint={2107.08430},
      archivePrefix={arXiv},
      primaryClass={cs.CV},
      url={https://arxiv.org/abs/2107.08430}, 
}

@article{hota,
  author    = {Jonathon Luiten and Aljo{\v{s}}a O{\v{s}}ep and Patrick Dendorfer and Philip Torr and Andreas Geiger and Laura Leal-Taixé and Bastian Leibe},
  title     = {HOTA: A Higher Order Metric for Evaluating Multi-object Tracking},
  journal   = {International Journal of Computer Vision},
  volume    = {129},
  number    = {2},
  pages     = {548--578},
  year      = {2021},
  doi       = {10.1007/s11263-020-01375-2},
  issn      = {1573-1405}
}

@article{ettrack,
  author    = {Xudong Han and Nobuyuki Oishi and Yueying Tian and Elif Ucurum and Rupert Young and Chris Chatwin and Philip Birch},
  title     = {ETTrack: enhanced temporal motion predictor for multi-object tracking},
  journal   = {Applied Intelligence},
  volume    = {55},
  number    = {1},
  pages     = {33},
  year      = {2024},
  url       = {https://doi.org/10.1007/s10489-024-05866-4},
  doi       = {10.1007/s10489-024-05866-4},
  issn      = {1573-7497}
}

@article{motiontrack,
    title = {MotionTrack: Learning motion predictor for multiple object tracking},
    journal = {Neural Networks},
    volume = {179},
    pages = {106539},
    year = {2024},
    issn = {0893-6080},
    doi = {https://doi.org/10.1016/j.neunet.2024.106539},
    author = {Changcheng Xiao and Qiong Cao and Yujie Zhong and Long Lan and Xiang Zhang and Zhigang Luo and Dacheng Tao}
}

@article{movesort,
  author    = {Momir Adžemović and Predrag Tadić and Andrija Petrović and Mladen Nikolić},
  title     = {Beyond Kalman filters: deep learning-based filters for improved object tracking},
  journal   = {Machine Vision and Applications},
  volume    = {36},
  number    = {1},
  pages     = {20},
  year      = {2024},
  doi       = {10.1007/s00138-024-01644-x},
  url       = {https://doi.org/10.1007/s00138-024-01644-x},
  issn      = {1432-1769}
}

@inproceedings{beehive_dataset,
author = {Bozek, Katarzyna and Hebert, Laetitia and Mikheyev, Alexander and Stephens, Greg},
year = {2018},
month = {06},
pages = {4185-4193},
title = {Towards Dense Object Tracking in a 2D Honeybee Hive},
doi = {10.1109/CVPR.2018.00440}
}

@INPROCEEDINGS{sort,
  author={Bewley, Alex and Ge, Zongyuan and Ott, Lionel and Ramos, Fabio and Upcroft, Ben},
  booktitle={2016 IEEE International Conference on Image Processing (ICIP)}, 
  title={Simple online and realtime tracking}, 
  year={2016},
  pages={3464-3468},
  doi={10.1109/ICIP.2016.7533003}
}

@inproceedings{deepsort,
  author={Wojke, Nicolai and Bewley, Alex and Paulus, Dietrich},
  booktitle={2017 IEEE International Conference on Image Processing (ICIP)}, 
  title={Simple online and realtime tracking with a deep association metric}, 
  year={2017},
  pages={3645-3649},
  doi={10.1109/ICIP.2017.8296962}
}

@inproceedings{linear_assignment,
  author = {Lyle Ramshaw and Robert E and Tarjan},
  title = {On Minimum-Cost Assignments in Unbalanced Bipartite Graphs},
  year = {2012},
  booktitle={HP Laboratories}
}

@article{infonce,
  title={Representation Learning with Contrastive Predictive Coding},
  author={Oord, Aaron van den and Li, Yazhe and Vinyals, Oriol},
  journal={arXiv preprint arXiv:1807.03748},
  year={2018}
}

@article{nuscenes,
  title={nuScenes: A multimodal dataset for autonomous driving},
  author={Holger Caesar and Varun Bankiti and Alex H. Lang and Sourabh Vora and 
          Venice Erin Liong and Qiang Xu and Anush Krishnan and Yu Pan and 
          Giancarlo Baldan and Oscar Beijbom},
  journal={arXiv preprint arXiv:1903.11027},
  year={2019}
}

@INPROCEEDINGS{retail_analytics,
  author={Hossam, Ahmed and Ramadan, Ahmed and Magdy, Mina and Abdelwahab, Raneem and Ashraf, Salma and Mohamed, Zeina},
  booktitle={2024 International Conference on Machine Intelligence and Smart Innovation (ICMISI)}, 
  title={Revolutionizing Retail Analytics: Advancing Inventory and Customer Insight with AI}, 
  year={2024},
  pages={64-69},
  doi={10.1109/ICMISI61517.2024.10580424}}

@ARTICLE{object_tracking_in_robotics,
  author={Xu, Zhefan and Zhan, Xiaoyang and Xiu, Yumeng and Suzuki, Christopher and Shimada, Kenji},
  journal={IEEE Robotics and Automation Letters}, 
  title={Onboard Dynamic-Object Detection and Tracking for Autonomous Robot Navigation With RGB-D Camera}, 
  year={2024},
  volume={9},
  number={1},
  pages={651-658},
  doi={10.1109/LRA.2023.3334683}}

@inproceedings{object_tracking_in_surveillence,
  author = {Oliver Urbann and Oliver Bredtmann and Maximilian Otten and Jan-Philip Richter and Thilo Bauer and David Zibriczky},
  title = {Online and Real-Time Tracking in a Surveillance Scenario},
  year = {2021},
  booktitle={5th Workshop on Long-term Human Motion Prediction},
  note = {Arxiv: 2106.01153}
}

@misc{mot_survey2025,
      title={Deep Learning-Based Multi-Object Tracking: A Comprehensive Survey from Foundations to State-of-the-Art}, 
      author={Momir Adžemović},
      year={2025},
      eprint={2506.13457},
      archivePrefix={arXiv},
      primaryClass={cs.CV},
      url={https://arxiv.org/abs/2506.13457}, 
}

@InProceedings{mpntrack,
    author={Guillem Brasó and Laura Leal-Taixé},
    title={Learning a Neural Solver for Multiple Object Tracking},
    booktitle = {The IEEE Conference on Computer Vision and Pattern Recognition (CVPR)},
    month = {June},
    year = {2020}
}

@InProceedings{sushi,
    author    = {Cetintas, Orcun and Bras\'o, Guillem and Leal-Taix\'e, Laura},
    title     = {Unifying Short and Long-Term Tracking With Graph Hierarchies},
    booktitle = {Proceedings of the IEEE/CVF Conference on Computer Vision and Pattern Recognition (CVPR)},
    month     = {June},
    year      = {2023},
    pages     = {22877-22887}
}

@article{simtrack,
  title={Backbone is All Your Need: A Simplified Architecture for Visual Object Tracking},
  author={Chen, Boyu and Li, Peixia and Bai, Lei and Qiao, Lei and Shen, Qiuhong and Li, Bo and Gan, Weihao and Wu, Wei and Ouyang, Wanli},
  journal={arXiv preprint arXiv:2203.05328},
  year={2022}
}

@inproceedings{sparsetrack,
  title={Sparsetrack: Multi-object tracking by performing scene decomposition based on pseudo-depth},
  author={Liu, Zelin and Wang, Xinggang and Wang, Cheng and Liu, Wenyu and Bai, Xiang},
  journal={IEEE Transactions on Circuits and Systems for Video Technology},
  year={2025},
  doi={10.1109/TCSVT.2024.3524670}
}

@article{ucmctrack,
  title={UCMCTrack: Multi-Object Tracking with Uniform Camera Motion Compensation}, 
  volume={38}, 
  DOI={10.1609/aaai.v38i7.28493}, 
  number={7}, 
  journal={Proceedings of the AAAI Conference on Artificial Intelligence}, 
  author={Yi, Kefu and Luo, Kai and Luo, Xiaolei and Huang, Jiangui and Wu, Hao and Hu, Rongdong and Hao, Wei},
  year={2024}, 
  month={Mar.}, 
  pages={6702-6710}
}

@article{boostrack,
  author    = {Vukasin D. Stanojevic and Branimir T. Todorovic},
  title     = {BoostTrack: boosting the similarity measure and detection confidence for improved multiple object tracking},
  journal   = {Machine Vision and Applications},
  volume    = {35},
  number    = {3},
  pages     = {53},
  year      = {2024},
  month     = {April},
  doi       = {10.1007/s00138-024-01531-5},
  url       = {https://doi.org/10.1007/s00138-024-01531-5},
  issn      = {1432-1769}
}

@INPROCEEDINGS{memot,
  author={Cai, Jiarui and Xu, Mingze and Li, Wei and Xiong, Yuanjun and Xia, Wei and Tu, Zhuowen and Soatto, Stefano},
  booktitle={2022 IEEE/CVF Conference on Computer Vision and Pattern Recognition (CVPR)}, 
  title={MeMOT: Multi-Object Tracking with Memory}, 
  year={2022},
  volume={},
  number={},
  pages={8080-8090},
  doi={10.1109/CVPR52688.2022.00792}
}

@article{twix,
    title = {Learning data association for multi-object tracking using only coordinates},
    journal = {Pattern Recognition},
    volume = {160},
    pages = {111169},
    year = {2025},
    issn = {0031-3203},
    doi = {https://doi.org/10.1016/j.patcog.2024.111169},
    url = {https://www.sciencedirect.com/science/article/pii/S0031320324009208},
    author = {Mehdi Miah and Guillaume-Alexandre Bilodeau and Nicolas Saunier}
}

@article{silu,
title = {Sigmoid-weighted linear units for neural network function approximation in reinforcement learning},
journal = {Neural Networks},
volume = {107},
pages = {3-11},
year = {2018},
note = {Special issue on deep reinforcement learning},
issn = {0893-6080},
doi = {https://doi.org/10.1016/j.neunet.2017.12.012},
author = {Stefan Elfwing and Eiji Uchibe and Kenji Doya}
}

@inproceedings{layernorm,
  title={Layer Normalization},
  author={Jimmy Lei Ba and Jamie Ryan Kiros and Geoffrey E. Hinton},
  year={2016},
  booktitle={arXiv preprint},
  note = {Arxiv: 1607.06450}
}

@article{dropout,
  title   = {Dropout: A Simple Way to Prevent Neural Networks from Overfitting},
  author  = {Srivastava, Nitish and Hinton, Geoffrey and Krizhevsky, Alex and Sutskever, Ilya and Salakhutdinov, Ruslan},
  journal = {Journal of Machine Learning Research},
  volume  = {15},
  number  = {56},
  pages   = {1929--1958},
  year    = {2014}
}

@inproceedings{adamw,
  title={Decoupled Weight Decay Regularization},
  author={Ilya Loshchilov and Frank Hutter},
  year={2019},
  booktitle={7th International Conference on Learning Representations (ICLR)}
}

@inproceedings{adam,
  title={Adam: A Method for Stochastic Optimization},
  author={Diederik P and Kingma and Jimmy Ba},
  year={2015},
  booktitle={3rd International Conference on Learning Representations (ICLR)}
}

@article{cosan,
  title   = {SGDR: Stochastic Gradient Descent with Warm Restarts},
  author  = {Loshchilov, Ilya and Hutter, Frank},
  journal = {arXiv preprint arXiv:1608.03983},
  year    = {2017}
}

@INPROCEEDINGS{ghost,
  author={Seidenschwarz, Jenny and Brasó, Guillem and Serrano, Victor Castro and Elezi, Ismail and Leal-Taixé, Laura},
  booktitle={2023 IEEE/CVF Conference on Computer Vision and Pattern Recognition (CVPR)}, 
  title={Simple Cues Lead to a Strong Multi-Object Tracker}, 
  year={2023},
  volume={},
  number={},
  pages={13813-13823},
  doi={10.1109/CVPR52729.2023.01327}}

@article{fairmot,
   title={FairMOT: On the Fairness of Detection and Re-identification in Multiple Object Tracking},
   volume={129},
   ISSN={1573-1405},
   DOI={10.1007/s11263-021-01513-4},
   number={11},
   journal={International Journal of Computer Vision},
   publisher={Springer Science and Business Media LLC},
   author={Zhang, Yifu and Wang, Chunyu and Wang, Xinggang and Zeng, Wenjun and Liu, Wenyu},
   year={2021},
   month=sep, pages={3069–3087} }

@article{putr,
  title   = {Is a Pure Transformer Effective for Separated and Online Multi-Object Tracking?},
  author  = {Liu, Chang and Zhan, Yao and Wu, Baigen and Zhang, Yang and Li, Wen and Wang, Wen and Tao, Dacheng},
  journal = {arXiv preprint arXiv:2405.14119},
  year    = {2024}
}

\end{document}